\documentclass{article}

\usepackage{microtype}
\usepackage{graphicx}
\usepackage{subcaption}
\usepackage{booktabs} % for professional tables

\usepackage{hyperref}

\usepackage[utf8]{inputenc}
\usepackage{amsmath, amssymb, amsthm}
\usepackage{graphicx}
\usepackage{hyperref}
\usepackage{amsmath}
\usepackage{natbib}
\usepackage{xspace}
\usepackage{listings}
\usepackage{relsize}
\usepackage{bm}
\usepackage{color}
\newcommand{\bx}{\bm{x}}
\newcommand{\bs}{\bm{s}}
\newcommand{\bv}{\bm{v}}

\newcommand{\methodShort}{GeoNorm\xspace}

\usepackage[preprint]{icml2026}

\usepackage{amsmath}
\usepackage{amssymb}
\usepackage{mathtools}
\usepackage{amsthm}

\usepackage[capitalize,noabbrev]{cleveref}

\theoremstyle{plain}

\theoremstyle{definition}

\theoremstyle{remark}

\usepackage[textsize=tiny]{todonotes}

\icmltitlerunning{GeoNorm: Unify Pre-Norm and Post-Norm with Geodesic Optimization}

\begin{document}

\twocolumn[
  \icmltitle{GeoNorm: Unify Pre-Norm and Post-Norm with Geodesic Optimization}



  \icmlsetsymbol{equal}{*}

  \begin{icmlauthorlist}
    \icmlauthor{Chuanyang Zheng}{1}
    \icmlauthor{Jiankai Sun}{2}
    \icmlauthor{Yihang Gao}{3}
    \icmlauthor{ Chi Wang}{4}
    \icmlauthor{ Yuehao Wang}{5}
    \icmlauthor{Jing Xiong}{6}
    \icmlauthor{ Liliang Ren}{7}
    \icmlauthor{ Bo Peng}{4}
    \icmlauthor{Qingmei Wang}{8}
    \icmlauthor{Xiaoran Shang}{9}
    \icmlauthor{ Mac Schwager}{2}
    \icmlauthor{Anderson Schneider}{1}
    \icmlauthor{Yuriy Nevmyvaka}{1}
    \icmlauthor{Xiaodong Liu }{7}
  \end{icmlauthorlist}

  \icmlaffiliation{1}{Morgan Stanley}
  \icmlaffiliation{2}{Stanford}
  \icmlaffiliation{3}{NUS}
  \icmlaffiliation{4}{Google}
  \icmlaffiliation{5}{UTA}
  \icmlaffiliation{6}{HKU}
  \icmlaffiliation{7}{Microsoft}
  \icmlaffiliation{8}{RUC}
  \icmlaffiliation{9}{USTC}

  \icmlcorrespondingauthor{Chuanyang Zheng}{cyzhengme@gmail.com}

  \icmlkeywords{Machine Learning, ICML}

  \vskip 0.3in
]



\printAffiliationsAndNotice{}  % no special notice (required even if empty)

\begin{abstract}
The placement of normalization layers, specifically Pre-Norm and Post-Norm, remains an open question in Transformer architecture design. In this work, we rethink these approaches through the lens of manifold optimization, interpreting the outputs of the Feed-Forward Network (FFN) and attention layers as update directions in optimization. Building on this perspective, we introduce GeoNorm, a novel method that replaces standard normalization with geodesic updates on the manifold. Furthermore, analogous to learning rate schedules, we propose a layer-wise update decay for the FFN and attention components. Comprehensive experiments demonstrate that GeoNorm consistently outperforms existing normalization methods in Transformer models. Crucially, GeoNorm can be seamlessly integrated into standard Transformer architectures, achieving performance improvements with negligible additional computational cost.
\end{abstract}

\section{Introduction}
Recent years have witnessed remarkable progress in Large Language Models (LLMs) \citep{brown2020language,ouyang2022training,touvron2023llama}, driven primarily by the exponential growth of training data and model parameters. This advancement is reflected in substantial improvements across language modeling \citep{fedus2022switch,puigcerver2023sparse,jiang2024mixtral,meta2025llama,liu2024deepseek,team2025kimi} and computer vision \citep{riquelme2021scaling,lin2023video}. Within the Transformer architecture, the choice of normalization function, such as Pre-Norm and Post-Norm, is critical, as it influences training stability, model performance, and gradient flow.

Currently, Pre-Norm has become the predominant choice in modern LLMs due to its favorable stability properties \citep{liu2024deepseek,yang2025qwen3,jiang2024mixtral}. In contrast, Post-Norm \citep{popel2018training,shazeer2018adafactor}, which applies normalization after the addition of the residual connection, was originally introduced in the Transformer and facilitates the training of deeper networks. However, it is often prone to training instability and loss spikes \citep{wang2024deepnet}. Pre-Norm was subsequently proposed to improve stability by applying layer normalization to the residual branch before the addition operation \citep{nguyen2019transformers}. Nonetheless, Pre-Norm introduces imbalanced gradients, with lower layers tending to receive larger updates than higher ones \citep{wang2024deepnet}.
To address these limitations, recent methods have sought to enhance stability and enable extreme model scaling. DeepNorm \citep{wang2024deepnet} stabilizes training and improves performance through better parameter initialization and residual scaling. Meanwhile, SandwichNorm \citep{ding2021cogview,yin2025pangu} introduces additional normalization layers at the network’s input and output. While these approaches demonstrate strong empirical results, they generally lack a unified theoretical framework for analyzing normalization functions in Transformers.

% normalization before and after the Feed-Forward Network and Attention Network.  In order to alleviate the above issue, there have been efforts to improve the optimization of deep Transformer by means of better initialization (Zhang et al., 2019a;b; Huang et al., 2020), or better architecture (Wang et al., 2019; Liu et al., 2020; Bachlechner et al., 2020; Shleifer et al., 2021). These approaches can stabilize a Transformer model with up to hundreds of layers. Yet, none of previous methods has been successfully scaled to 1,000 layers.

% Pre-Norm and Post-Norm are commonly used in Transformer, while there is still no enough exploration. Current,y the Pre-Norm is widnely used in Large Language model because of the easy and relatively stable training. The Post-Norm is suggested to have better performance with more layers. The DeepNorm proposed to change the weight of input and residual for stable training. And the Standwichnorm propose to add normalization before and after the feed-forward network and attention network. However, all this works do not build a theoretical analysis what is the meaning of the $x+ffn(x)$  and $x_{attention}(x)$. In this work, we explain the $x+ffn(x)$  and $x_{attention}(x)$ as the optimization on the sphere, while the $x$ is the current point, and $ffn(x)$ and $attention(x)$ is the estimated gradient.

In this work, we interpret the operations \(\bx + \mathrm{FFN}(\bx)\) and \(x + \mathrm{Attention}(\bx)\) as optimization steps on a sphere, where \(\mathrm{FFN}(\bx)\) and \(\mathrm{Attention}(\bx)\) serve as estimated gradient directions produced by the respective modules. From this perspective, each Transformer layer can be viewed as performing an iterative optimization step within a dynamical system: token embeddings act as the initial state, while the attention and feed-forward modules generate update directions based on the current iterate. The normalization operator then projects the updated vector back onto a constraint set, specifically, the sphere defined by the normalization radius.
Building on this formulation, our contributions are threefold:

\begin{itemize}
    \item \textbf{Theoretical Framework:} We introduce a theoretical framework for analyzing normalization in Transformers. Here, \(\bx\) is treated as the current point, while \(\mathrm{FFN}(\bx)\) and \(\mathrm{Attention}(\bx)\) are interpreted as the update direction in optimization.

    \item \textbf{\methodShort\ Normalization:} We propose \methodShort, a novel normalization function that replaces conventional schemes by leveraging geodesic and Riemannian optimization on the sphere, leading to improved performance.

    \item \textbf{Extensive Empirical Validation:} Through comprehensive experiments across varying training lengths, datasets, and model sizes, we validate the effectiveness of \methodShort. Additionally, we demonstrate its scalability and generalization on large-scale datasets with downstream task evaluation.
\end{itemize}

\section{Related Work}

\paragraph{Normalization Function} The original Post-Norm architecture, first introduced in the Transformer model \cite{popel2018training, shazeer2018adafactor}, applies layer normalization to the sum of the identity term and the residual output. While Post-Norm facilitates the training of deep networks, it can suffer from instability and exhibit sharp loss spikes \cite{wang2024deepnet}. In response, the Pre-Norm variant \cite{nguyen2019transformers} modifies the residual connection by applying layer normalization directly to the identity term before the addition, leading to improved training stability. However, this often results in imbalanced gradient magnitudes, with lower layers receiving larger gradients compared to higher layers \cite{wang2024deepnet}.
To further enhance stability and enable scaling to significantly greater depths, DeepNorm \cite{wang2024deepnet} introduces refined parameter initialization and residual scaling. Another approach, StandwichNorm \cite{ding2021cogview,yin2025pangu}, incorporates normalization layers at both the network input and output alongside the identity connection. Although these methods have empirically demonstrated strong performance, they currently lack a comprehensive theoretical framework for analyzing the underlying normalization mechanisms.

\paragraph{Gradient Descent.} The stochastic gradient descent (SGD) \cite{bottou1998online} is an iterative optimization method with the gradient. Based on the SGD, there is SGD with momentum \cite{rumelhart1986learning} which use the momentum of previous gradient.  Adagrad (adaptive gradient algorithm) \cite{duchi2011adaptive} is a modified stochastic gradient descent algorithm with per-parameter learning rate. The RMSProp \cite{graves2013generating} divide the learning rate for a weight by a running average of the magnitudes of recent gradients for that weight. Adam \cite{kingma2017adammethodstochasticoptimization} run averages with exponential forgetting of both the gradients and the second moments of the gradients are used.

\paragraph{Optimization Method.} The Geodesic optimization is related to the optimization on the manifold \cite{kingma2017adammethodstochasticoptimization}. Riemannian Steepest Descent Method extends the classic gradient descent to optimize functions on curved spaces (manifolds) \cite{debye1909naherungsformeln} by following the path of steepest decrease, using the Riemannian gradient (the projection of the standard gradient onto the manifold) and a retraction to map steps back onto the manifold.  Riemannian Newton's method \cite{da2025thorough} adapts classical Newton's method for optimization problems on curved spaces (Riemannian manifolds).   The Riemannian Conjugate Gradient method \cite{hestenes1952methods} extends the conjugate gradient method to a manifold. For the retraction relation method, they will calculate the gradient and map the result back to the manifold. The geodesic-based methods \cite{zhang2016first} use the gradient descent method on the geodesic of the manifold.

\section{Method}

\subsection{Post-Normalization in LLMs}
Beyond the well-recognized attention and feed-forward network modules, modern LLMs rely critically on normalization to stabilize training and inference as depth increases.
Let $\bx_{k}$ denote the token representation after the $k$-th Transformer layer.
A Transformer layer with post-normalization can be written as
\begin{equation}
\label{eq1}
    \begin{split}
        & \Tilde{\bx}_{k} = \text{Norm}(\bx_{k} + \text{Attn}(\bx_k)),\\
        & \bx_{k+1} = \text{Norm}(\Tilde{\bx}_{k}  + \text{FFN}(\Tilde{\bx}_{k})),
    \end{split}
\end{equation}
where $\text{Attn}(\cdot)$ and $\text{FFN}(\cdot)$ denote the attention and feed-forward modules, and $\text{Norm}(\cdot)$ is the normalization operator (typically RMSNorm followed by rescaling to preserve the vector norm).

From this formulation, we may view the Transformer layer as performing an iterative optimization step in a dynamical system. The token embeddings serve as the initial state, while the attention and feed-forward modules produce update directions based on the current iterate $\bx_k$. 
The normalization operator then projects the updated vector back to a constraint set, specifically, the sphere defined by the normalization radius $\|\bx_k\|$. This interpretation closely mirrors the projected optimization scheme:
\begin{equation}
\label{eq2}
    \bx_{k+1} = \text{Proj}_{\bm{\Omega}}(\bx_{k} + \alpha_k \bs_k),
\end{equation}
where $\bs_k$ is an update step determined by past iterates $\{\bx_{0},\bx_{1},\cdots,\bx_{k}\}$, $\alpha_k$ represents the step size, and $\text{Proj}_{\bm{\Omega}}(\cdot)$ denotes projection onto the feasible region $\bm{\Omega}$. Equation \eqref{eq2} includes a wide range of classical optimization algorithms such as projected gradient descent.

From this perspective, \eqref{eq1} corresponds to the special case where the feasible region $\bm{\Omega}$ is the sphere in the Euclidean space induced by $\ell_2$-norm, and the projection $\text{Proj}_{\bm{\Omega}}(\cdot)$ is simply vector normalization.
Thus, each Transformer layer applies an unconstrained update followed by a projection back to the sphere, aligning post-normalization Transformers with the general class of optimization algorithms that incorporate projection to enforce constraints. This connection motivates replacing the projection step with a more geometrically informed operation, such as the exponential mapping, in the subsequent section.

\subsection{Geodesic-Inspired Transformer Design}

Geodesic-based optimization methods are known to possess several structural advantages over projected gradient approaches, particularly when the feasible region $\bm{\Omega}$ is a smooth manifold (e.g., spheres, Stiefel/orthogonal matrices, low-rank matrices, or fixed-rank positive semi-definite matrices).
Traditional projected methods handle constraints by alternating between Euclidean optimization and projection back onto the feasible set. While simple to implement and interpret, this extrinsic approach discards geometric information that the projection step can significantly distort step $\bs_k$, introduce unexpected kinks in the optimization trajectory, and degrade convergence due to repeated violations and corrections of feasibility. 
These issues are amplified when the feasible region is a smooth manifold, where the curvature causes projected steps to deviate substantially from the true directions. 
In contrast, geodesic-based optimization circumvents these limitations by performing updates intrinsically on the manifold. By following geodesics or retraction paths, each iterate remains feasible without projection, and the search direction corresponds to the Riemannian step under the manifold metric. 
As a consequence, geodesic-based methods typically exhibit smoother optimization trajectories, improved conditioning, and stronger convergence guarantees compared to their projected counterparts.
These observations motivate our modification of the Transformer model from projection-based schemes in \eqref{eq1} to a geodesic-based formulation.

In geodesic optimization, the core idea is to modify the step direction $\bs_k$ in a manner consistent with the geometry of the manifold $\bm{\Omega}$, ensuring that every iterate remains on $\bm{\Omega}$.
This is achieved through the exponential map $\text{exp}_{\bx}(\cdot)$. 
Consider a smooth manifold $\bm{\Omega}$ with tangent space $T_{\bx}\bm{\Omega}:= \{\bv: \bx^{\top}\bv = 0\}$ at point $\bx$. 
For any tangent vector $\bv \in T_{\bx}\bm{\Omega}$, there exists a unique geodesic $\gamma_{\bv}(t)$ satisfying 
\begin{equation*}
    \gamma_{\bv}(0) = \bx,  \quad \gamma_{\bv}^{\prime}(0) = \bv.
\end{equation*}
The exponential map at $\bx$ is then defined as $\text{exp}_{\bx}(\bv) = \gamma_{\bv}(1)$. We refer readers to  Riemannian optimization literatures~\cite{zhang2016first,da2025thorough} for formal definitions and properties of exponential mappings. 
For a general manifold $\bm{\Omega}$, the exponential map is implicit and rarely available in a closed form, which limits the practical adoption of geodesic-based methods due to the computational cost of evaluating $\text{exp}_{\bx}(\bv)$.
However, in our setting \eqref{eq1}, the manifold is the sphere, where the exponential map is explicit and computationally efficient.

Let $\bm{\Omega}$ be a sphere embedded in Euclidean space with the standard induced Riemannian metric. Given a point $\bx \in \bm{\Omega}$ and a tangent vector $\bv \in T_{\bx}\bm{\Omega}$, the exponential map admits the closed-form expression
\begin{equation}
\label{eq3}
    \text{exp}_{\bx}(\bv) = \cos\left(\frac{\|\bv\|}{\|\bx\|}\right) \bx + \|\bx\| \sin\left(\frac{\|\bv\|}{\|\bx\|}\right) \frac{\bv}{\|\bv\|}.
\end{equation}
Given an update direction $\bs$ for $\bx$, we first project it to the tangent space:
\begin{equation*}
    \bv = \bs - \frac{\bx^{\top}\bs}{\|\bx\|^2} \bx,
\end{equation*}
and then update the representation intrinsically via the exponential map $\bx + \text{exp}_{\bx}(\bv)$. Specifically, \eqref{eq2} is modified to
\begin{equation}
    \begin{split}
        & \bv_k = \bs_k - \frac{\bx_k^{\top}\bs_k}{\|\bx_k\|^2} \bx_k,\\
        & \bx_{k+1} = \text{exp}_{\bx_k}(\alpha_k \bv_k).
    \end{split}
\end{equation}
Exponential mapping interprets the update as moving along the geodesic determined by the update direction, producing a smooth, norm-preserving, and curvature-informed transformation between states.

These observations motivate us to replace the projection-based normalization in \eqref{eq1} with a geodesic-inspired normalization based on the exponential map \eqref{eq3}.
More specifically, we reformulate the post-normalized Transformer layer \eqref{eq1} as
\begin{equation}
\label{eq4}
    \begin{split}
        & \bs_{k} = \text{Attn}(\bx_k),\\
        & \bv_{k} = \bs_{k} - \frac{\bx_{k}^{\top} \bs_{k}}{\|\bx_{k}\|^2} \bx_{k},\\
        & \Tilde{\bx}_{k} = \text{exp}_{\bx_k}(\alpha_{k}\bv_{k}),\\
        & \Tilde{\bs}_{k} = \text{FFN}(\Tilde{\bx}_{k}),\\
        & \Tilde{\bv}_{k} = \Tilde{\bs}_{k} - \frac{\Tilde{\bx}_{k}^{\top} \Tilde{\bs}_{k}}{\|\Tilde{\bx}_{k}\|^2} \Tilde{\bx}_{k},\\
        & \bx_{k+1} = \text{exp}_{\Tilde{\bx}_k}(\alpha_{k}\Tilde{\bv}_{k}),
    \end{split}
\end{equation}
where the update directions $\bs_k$ and $\Tilde{\bs}_k$ are produced by the attention and feed-forward modules, $\bv_k$ and $\Tilde{\bv}_k$ are their tangent-space projections, and $\{\alpha_k\}$ is a prescribed sequence of step sizes. 
In contrast to the standard post-normalization, which abruptly projects the updated representation back to the sphere, the update in \eqref{eq4} moves intrinsically along the geodesic determined by the residual direction.

Step-size selection is an important component of geodesic optimization. Classical choices include constant step sizes for geodesically convex and smooth objectives ($\alpha_k \leq \frac{2}{L}$, where $L$ is the gradient Lipschitz constant), polynomially diminishing step sizes for stochastic Riemannian optimization, and Barzilai–Borwein (BB) step sizes for accelerated variants.
Since the token representations in Transformer layers evolve under stochastic transformations, arising from dropout, token generation sampling, autoregressive randomness, and the non-deterministic interaction of multiple tokens, we model the layerwise dynamics as a stochastic Riemannian optimization process.
Accordingly, we adopt a polynomially decaying step size of the form
$\alpha_k = \frac{\alpha}{k^{0.5}}$ (named as \textbf{sqrt}), which theoretically achieves the $O(\frac{\log k}{k})$ convergence rate for stochastic Riemannian gradient methods~\cite{da2025thorough}. Also we could also have  $\alpha_k = \frac{\alpha}{k}$ (named as \textbf{harmonic}) and $\alpha_k = \frac{\alpha (T-K)}{T}$ (named as \textbf{linear}, and $T$ is the total layer number).

In the proposed method \eqref{eq4}, we keep the attention and FFN modules unchanged and modify only the post-normalization step. Instead of applying a projection via the normalization operator $\text{Norm}(\cdot)$, we adopt the exponential map inspired by geodesic Riemannian optimization. Our intuition is that projection-based normalization may destroy certain geometric structure in the update directions $\bs_k$ and $\Tilde{\bs}_k$ generated by the attention and FFN modules. Projection forces the representation back onto the sphere by discarding the radial component abruptly, potentially altering the intended direction of motion. 
In contrast, the exponential map preserves the intrinsic geometry of the sphere: it transports the tangent vector faithfully along the geodesic without the directional distortions introduced by renormalization. This allows the model to follow the spherical manifold structure naturally, maintaining richer geometric relationships between the current representation $\bx_k$, the attention update $\bs_{k}$, the FFN update $\Tilde{\bs}_k$, and the curvature of the underlying feature space.
Comparing the classical Transformer update \eqref{eq1} with our geodesic formulation \eqref{eq4}, we note that the exponential map adds negligible computational overhead. Its closed-form expression involves only basic vector operations and introduces no additional trainable parameters. Consequently, our method integrates seamlessly into existing LLM architectures without modifying the core Transformer blocks, and is computationally efficient.

\subsection{Pre-Norm with GeoNorm formulation}
 Moreover, we find that Pre-Norm can be regarded as a special case of GeoNorm (as shown in Appendix \ref{appendix: prenorm}), as it follows an update scheme similar to \eqref{eq4}, with the corresponding mapping in \eqref{eq3} but with non-adaptive angles.
This interpretation is further supported by empirical results, where the two formulations exhibit comparable performance.

In summary, we interpret Transformers as Riemannian flows of token embeddings evolving on a spherical manifold. Replacing projection with geodesic-informed exponential updates in \eqref{eq4} enables representations to move smoothly along circles on the sphere, leading to more coherent and curvature-aware information propagation across layers.

\section{Experiment}
\paragraph{Baseline.}
We compare the proposed \methodShort with the $\mathrm{Dense}$ model and normalization.
To be specific, we evaluate \methodShort against normalization methods, including Pre-Norm, Post-Norm, DeepNorm and SandwichNorm. \textbf{The implementation of \methodShort is in Appendix \ref{appendix: implementation}}.

% \paragraph{Datasets}
% \vspace{-5pt}
\paragraph{Datasets.}
Our analysis involves training language models on the Arxiv and Books3 datasets, which are frequently used benchmarks for evaluating model performance. Moreover, we train the model on the large-scale dataset FinWeb-Edu \citep{lozhkov2024fineweb-edu} and evaluate on downstream datasets, 
% including ARC~\citep{clark2018think},  HellaSwag~\citep{zellers2019hellaswag}, OBQA \cite{mihaylov2018can}, SciQ \citep{welbl2017crowdsourcing}, PIQA \citep{bisk2020piqa},  WinoGrade~\citep{sakaguchi2021winogrande}, SocialIQA \cite{sap2019socialiqa} and BoolQ \cite{clark2019boolq}.
% % \vspace{-5pt}
\paragraph{Experiment settings.}
Initially, we compare \methodShort with other baselines at training lengths 512 and 1024, using decoder-only Transformers \citep{brown2020language} with model size 125M, whose configuration is shown in Appendix \ref{model configuration details}. Subsequently, we evaluate the performance of larger model sizes, specifically 350M and 1.3 B.

\subsection{Compare with baseline}
\begin{figure*}[htbp]
%\begin{figure*}
\centering
\includegraphics[width=0.45\textwidth]{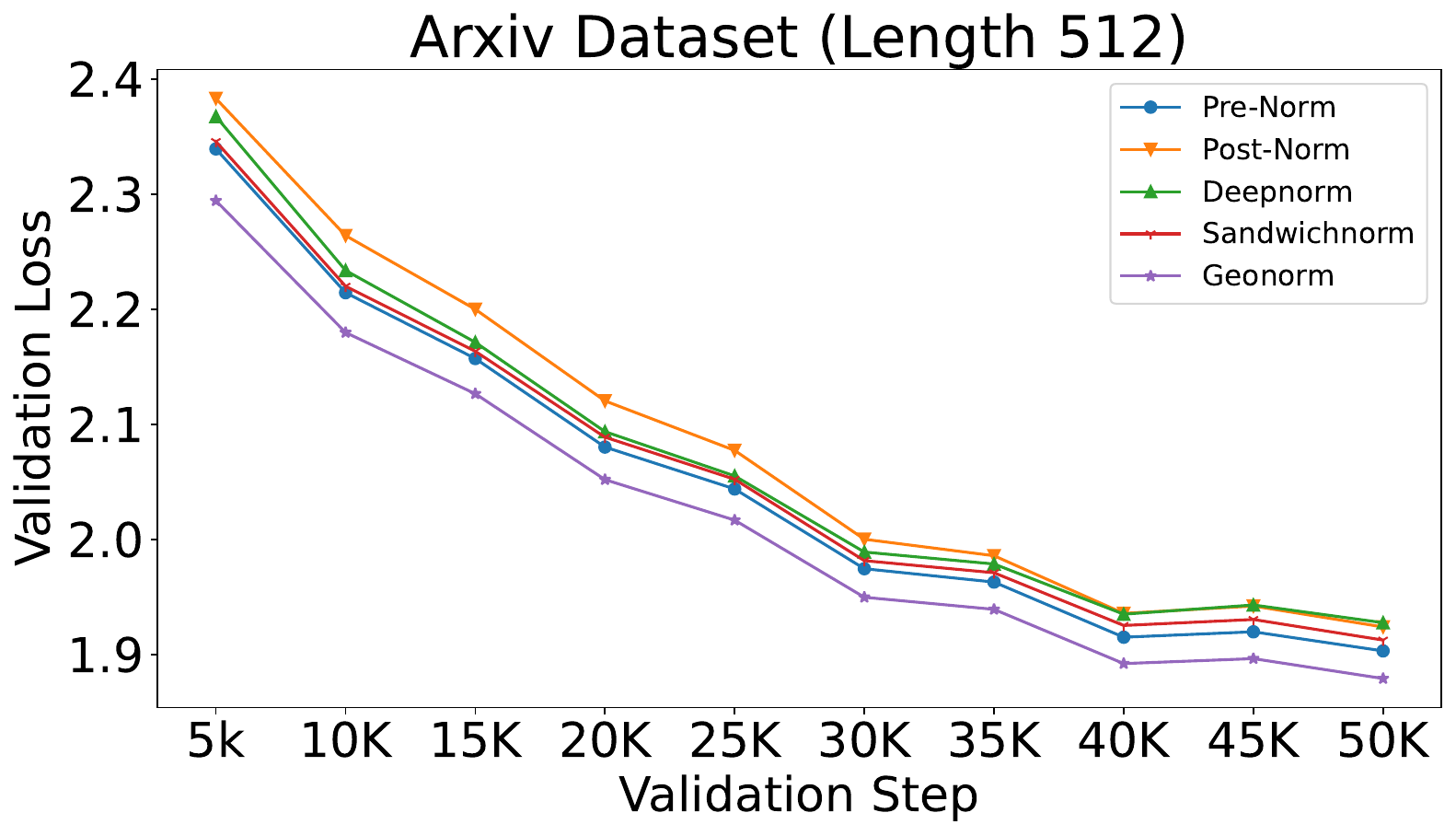}
\hspace{0in}
\includegraphics[width=0.45\textwidth]{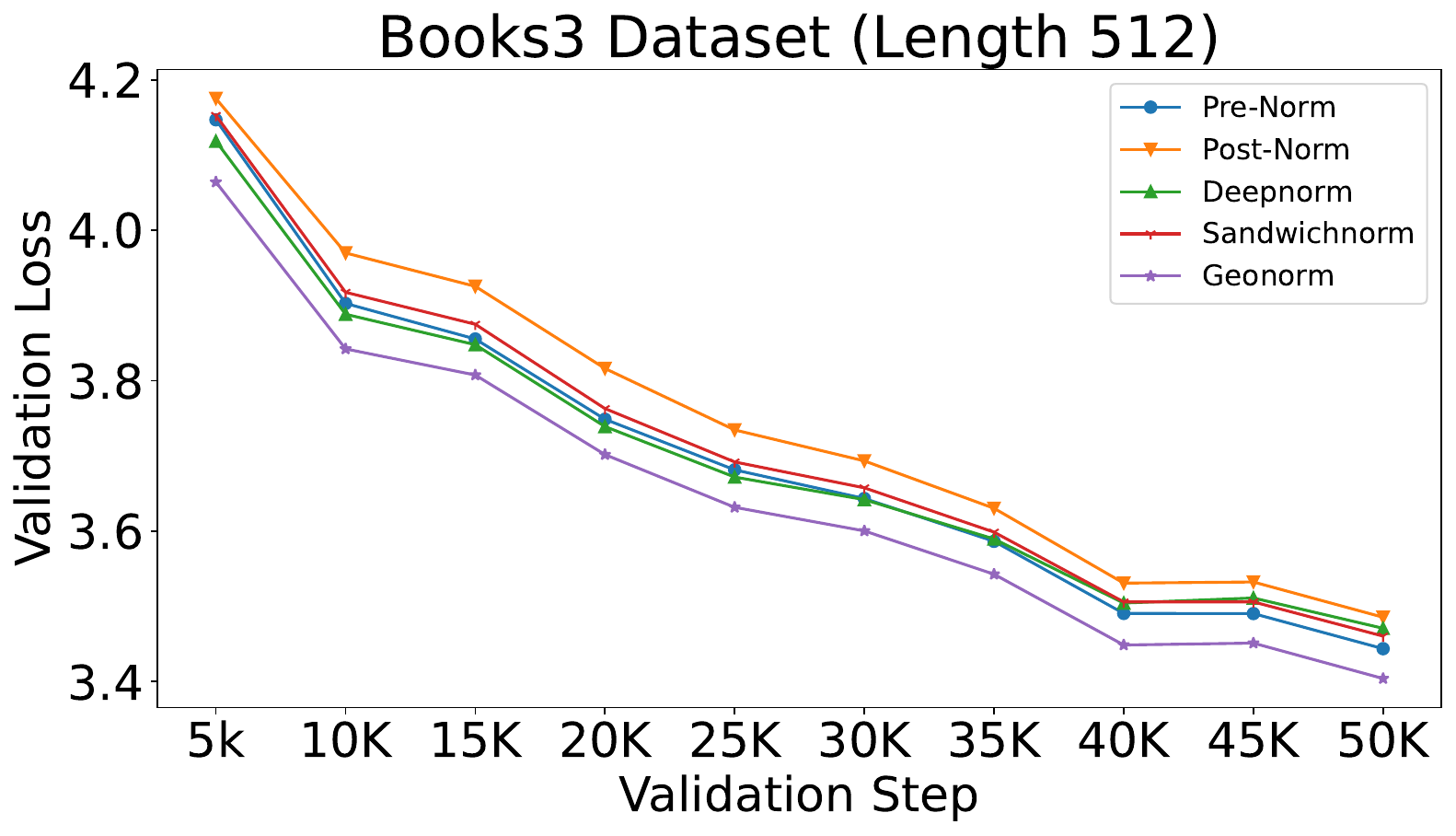}
\includegraphics[width=0.45\textwidth]{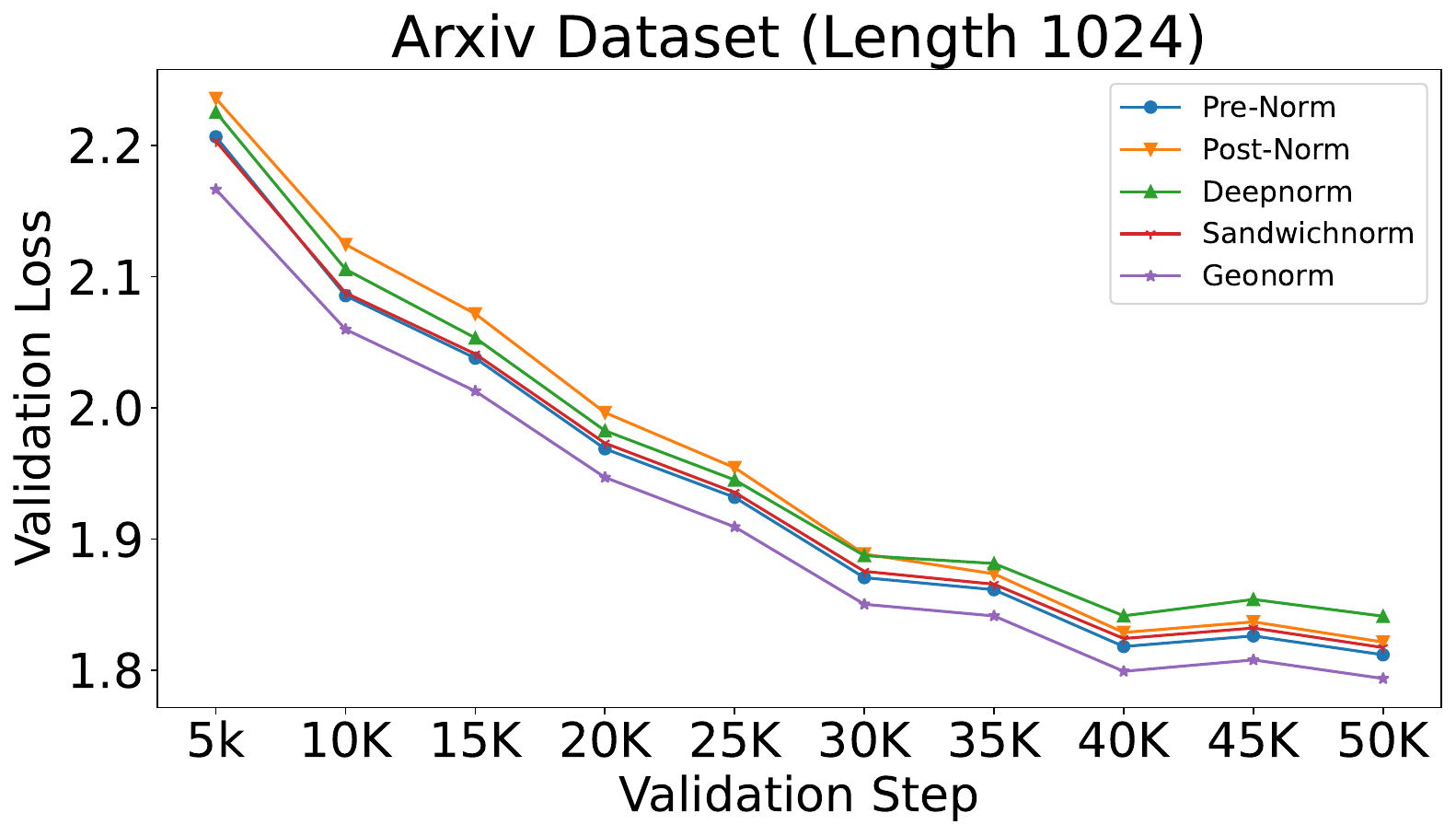}
\hspace{0in}
\includegraphics[width=0.45\textwidth]{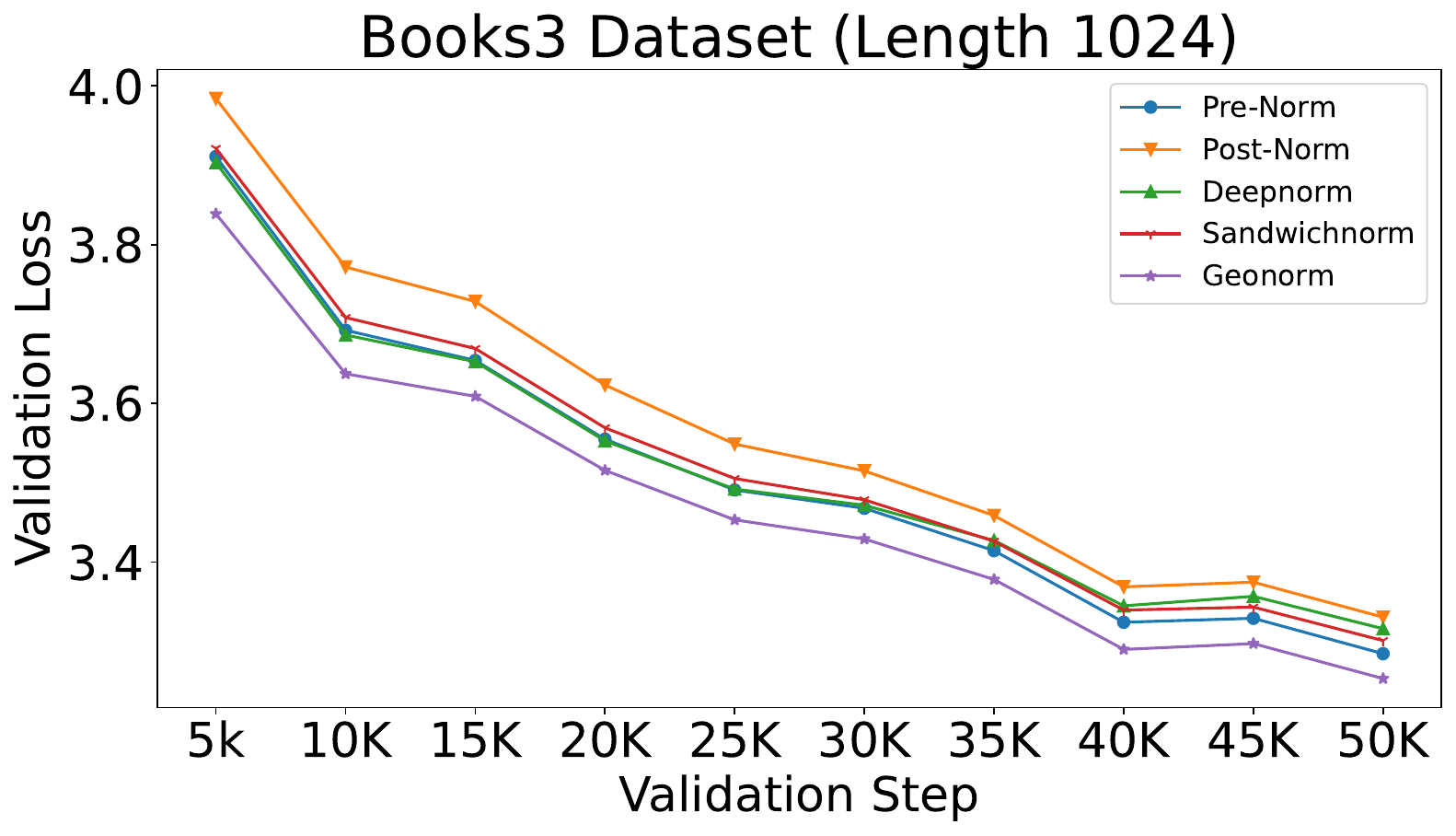}
\caption{
The performance of different methods on the Arxiv and Books3 dataset, with model parameter 125M.
}
% \vspace{-5pt}
\label{fig: compare with baseline}
\end{figure*}
\paragraph{\methodShort achieves better performance, with different datasets.} As shown in Figure \ref{fig: compare with baseline}, 
to evaluate the generalization capability of the proposed method, 
we conducted experiments across two distinct datasets, Arxiv and Books3, using a fixed training length of 512. The results demonstrate a clear and consistent performance hierarchy. 
On the Arxiv dataset, \methodShort establishes a strong lead with a final loss of 1.8792. This outperforms the closest baseline, Pre-Norm (1.9032), as well as Post-Norm (1.9240), DeepNorm (1.9277), and SandwichNorm (1.9126). Crucially, this pattern of superiority is not an artifact of a single data distribution. It is robustly replicated on the Books3 dataset, where \methodShort again achieves the optimal loss of 3.4040, maintaining a decisive margin over all alternative normalization methods. The fact that \methodShort secures the top position on both datasets, which represent different textual domains and complexities, provides compelling evidence that its architectural advantage is not dataset-specific. Therefore, we conclude that \methodShort offers generalizable improvements, delivering reliably better performance across diverse data environments.

\paragraph{\methodShort achieves better performance, with different training lengths.} The performance of \methodShort demonstrates consistent superiority across varying training lengths on the Books3 dataset. At a training length of 512, \methodShort establishes an immediate lead, achieving an optimal loss of 3.4040. This already surpasses the results of all comparative baselines, including Pre-Norm (3.4437), Post-Norm (3.4855), DeepNorm (3.4708), and SandwichNorm (3.4604). Crucially, this initial advantage is not merely maintained but is significantly amplified when the training duration is extended. At the longer training length of 1024, \methodShort again records the best possible loss of 3.2534. More importantly, the performance gap separating it from the nearest competitor widens considerably at this scale; while Pre-Norm achieves a loss of 3.2847, the other methods fall even further behind. 
% This pattern confirms two key findings: first, \methodShort delivers the best absolute performance at every tested training length; and second, its relative improvement over alternative methods scales positively with increased training compute.
Therefore, the superiority of \methodShort is both robust and generalizes effectively across key experimental parameters.

\paragraph{\methodShort is consistently better across the validation steps.} The evaluation of training dynamics reveals that \methodShort maintains a consistent performance advantage throughout the entire validation schedule. At the initial validation step (5000), \methodShort establishes an early lead with a loss of 4.0642, which is superior to all baseline methods. This advantage is not transient; from step 5000 through 50000, \methodShort continues to achieve the lowest loss at every measured interval, demonstrating stable and sustained optimization. 
% Crucially, this superior performance persists until the final validation step at the end of training. 
The fact that \methodShort leads at the beginning, maintains the lead throughout, and finishes with the best performance provides strong evidence that its architectural benefits are robust across all training phases. 
Therefore, we conclude that \methodShort consistently delivers better performance, showcasing reliable convergence and optimization stability.

\subsection{Performance on Large Model}
\begin{figure*}[htbp]
%\begin{figure*}
\centering
\includegraphics[width=0.45\textwidth]{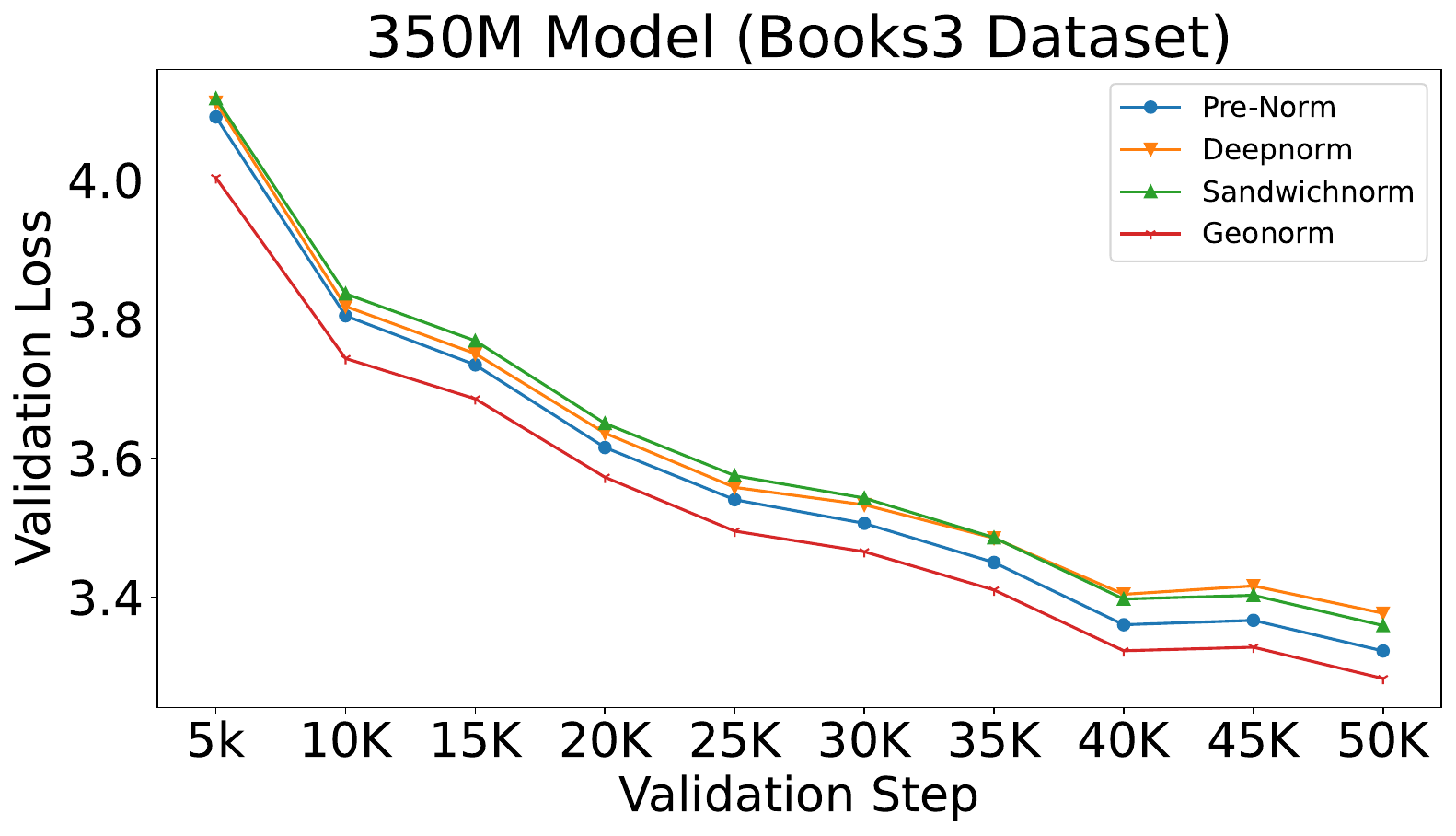}
\hspace{0in}
\includegraphics[width=0.45\textwidth]{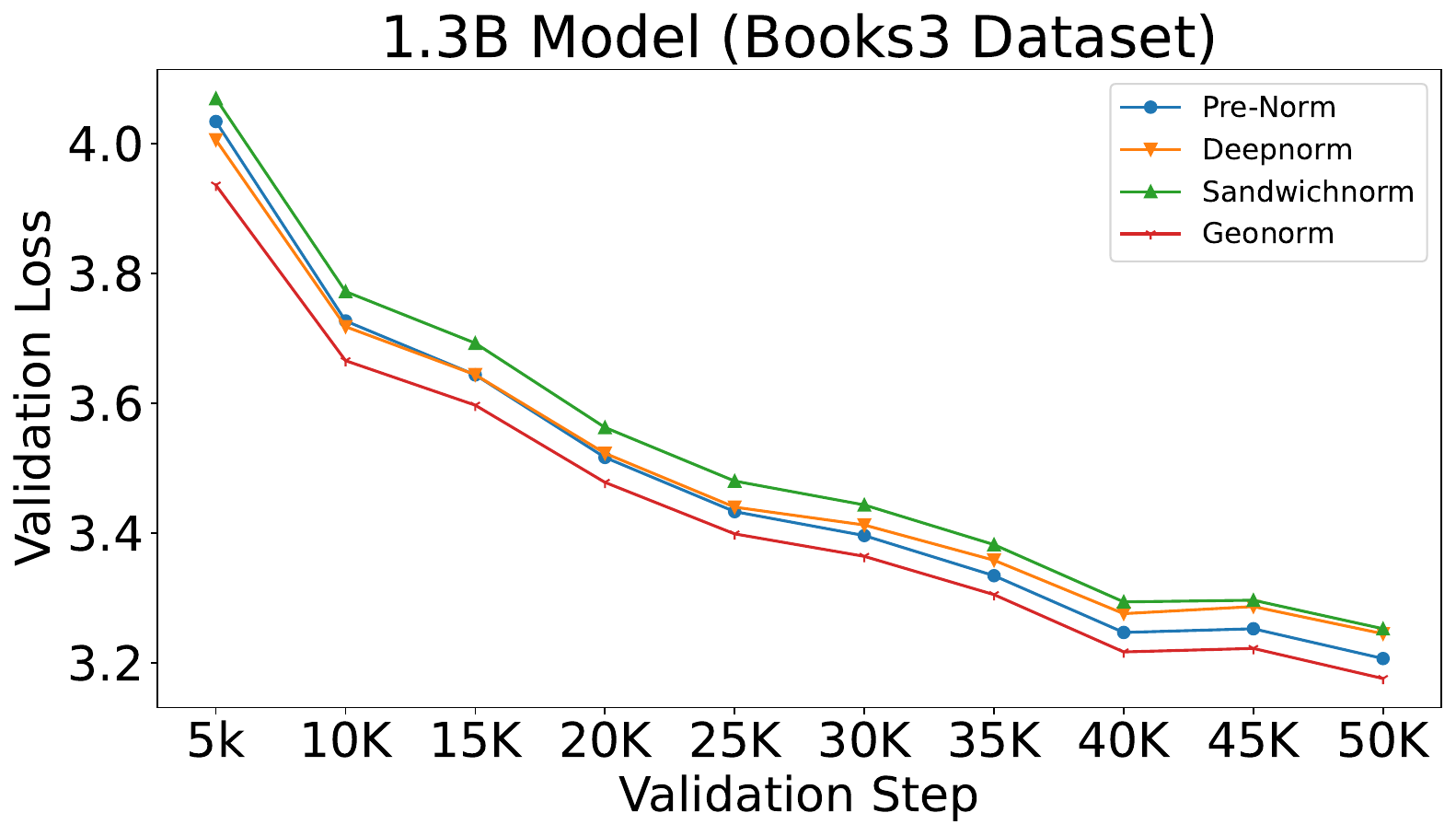}
\caption{
The performance of different baselines on the Books3 dataset, with model parameter 350M and 1.3B, training lengths of 512.
}
% \vspace{-5pt}
\label{fig: large model}
\end{figure*}

\paragraph{The \methodShort achieves the best performance with larger model size.} As shown in Figure \ref{fig: large model}, the performance of \methodShort demonstrates consistent superiority across increasing model sizes. For a 350M parameter model, it achieves the best loss of 3.2837, compared to 3.3233 (Pre-Norm), 3.3775 (DeepNorm), and 3.3598 (SandwichNorm). Scaling to 1.3B parameters, \methodShort again leads with a loss of 3.1759, while the next best baseline, Pre-Norm, achieves only 3.2067. These results confirm two key points: first, \methodShort delivers the best performance at every tested scale, and second, its performance advantage persists and even improves relative to alternatives as model size grows, showcasing its scalability.

\paragraph{\methodShort consistently achieves the best performance across the validation steps.} \methodShort also shows stable performance throughout training, maintaining the best validation loss at multiple checkpoints. At validation step 5000, the 350M \methodShort model achieves the lowest loss, and this advantage persists at step 50000. The same pattern holds for the 1.3B model, which outperforms baselines from early to late validation steps. These results confirm that \methodShort's performance advantages are robust across different training stages.

\subsection{Performance on Long Training Length}

\begin{figure*}[htbp]
%\begin{figure*}
\centering
\includegraphics[width=0.45\textwidth]{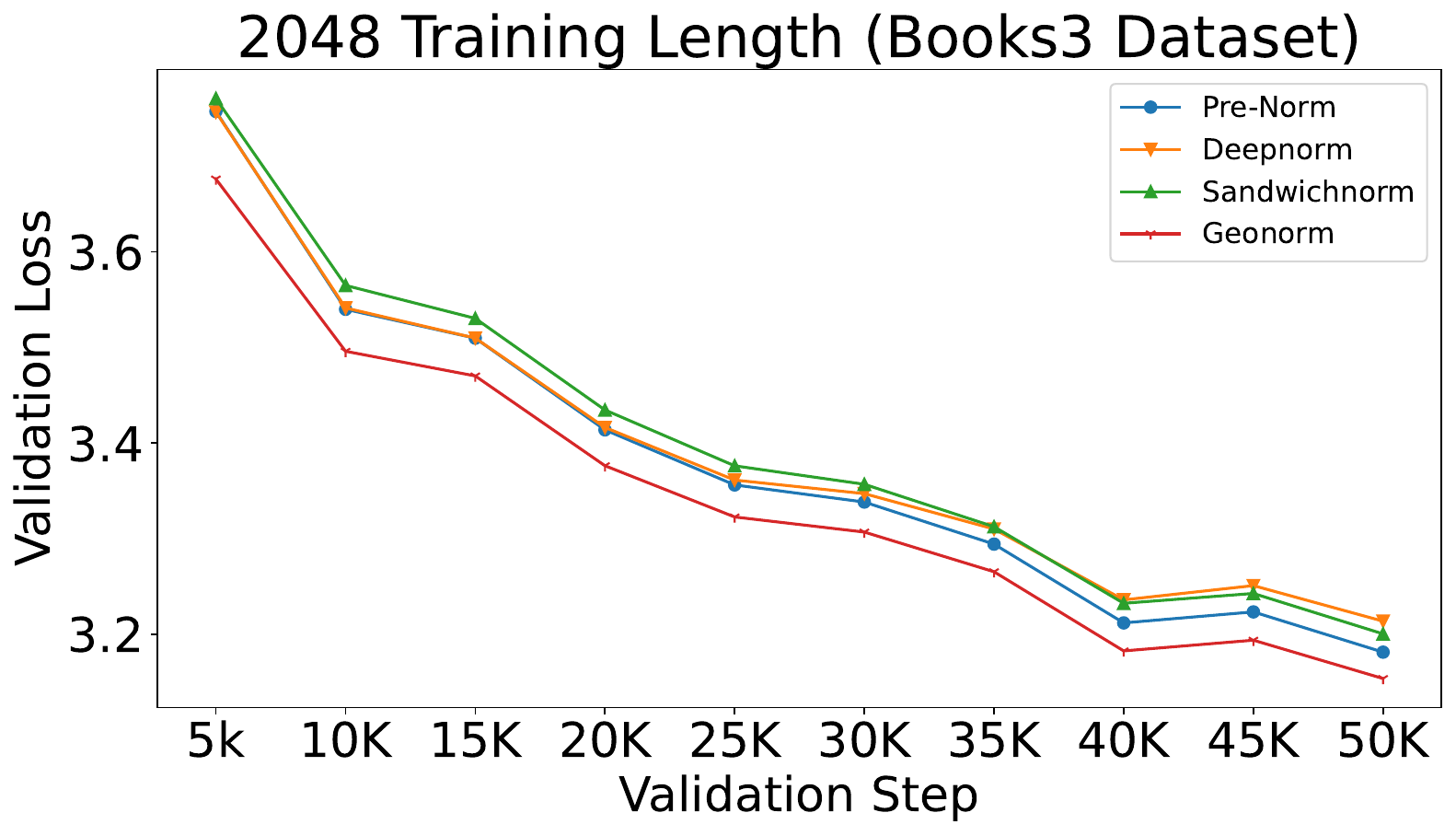}
\hspace{0in}
\includegraphics[width=0.45\textwidth]{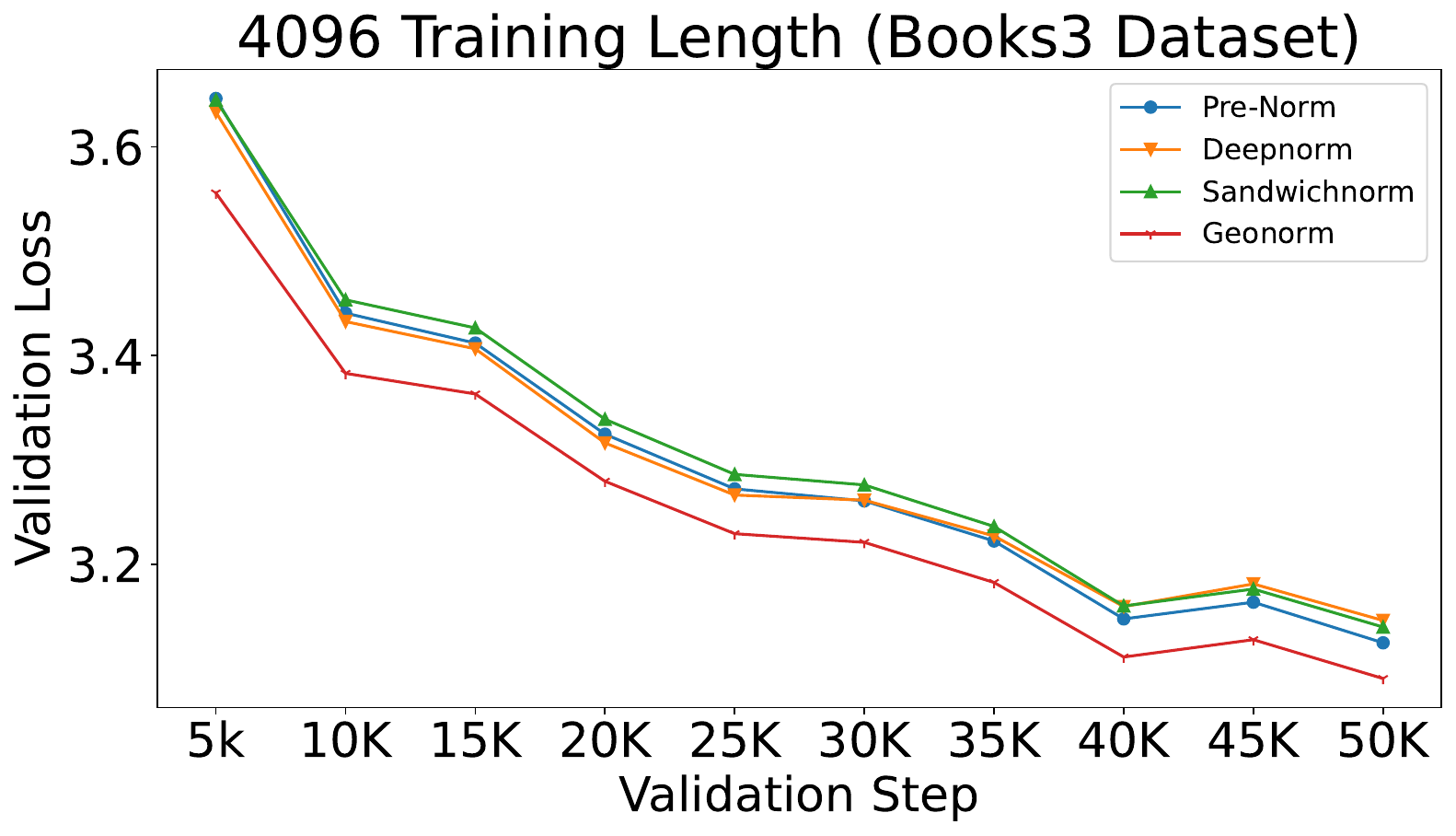}
\caption{
The performance of different baselines on the Books3 dataset, with model size 125M and training length 2048 and 4096.
}
% \vspace{-5pt}
\label{fig: long training length}
\end{figure*}

\paragraph{The \methodShort keeps the best performance with longer training length.} As shown in Figure \ref{fig: long training length}, our results demonstrate the consistent superiority of the \methodShort method across different training lengths. At a length of 2048, it achieves the best performance with a loss of 3.1538, outperforming Pre-Norm, DeepNorm, and SandwichNorm. Extending the training length to 4096 further improves its performance to a loss of 3.0908, while maintaining its lead over all baseline methods. This dual finding—performance improvement with extended training and sustained superiority—confirms that \methodShort not only provides the best initial results but also retains its optimal performance advantage as training progresses.

\paragraph{The training of \methodShort is stable, while others may have a loss spike.} With a training length of 4096, the post-norm presents the loss spike and ruins the model training. On the other hand, the \methodShort training is relatively stable. Therefore, \methodShort training is relatively stable, while others (e.g. Post-Norm) may have a loss spike and the training process will collapse.

\subsection{The Effect of decay method}
\begin{figure*}[!htbp]
%\begin{figure*}
\centering
\includegraphics[width=0.45\textwidth]{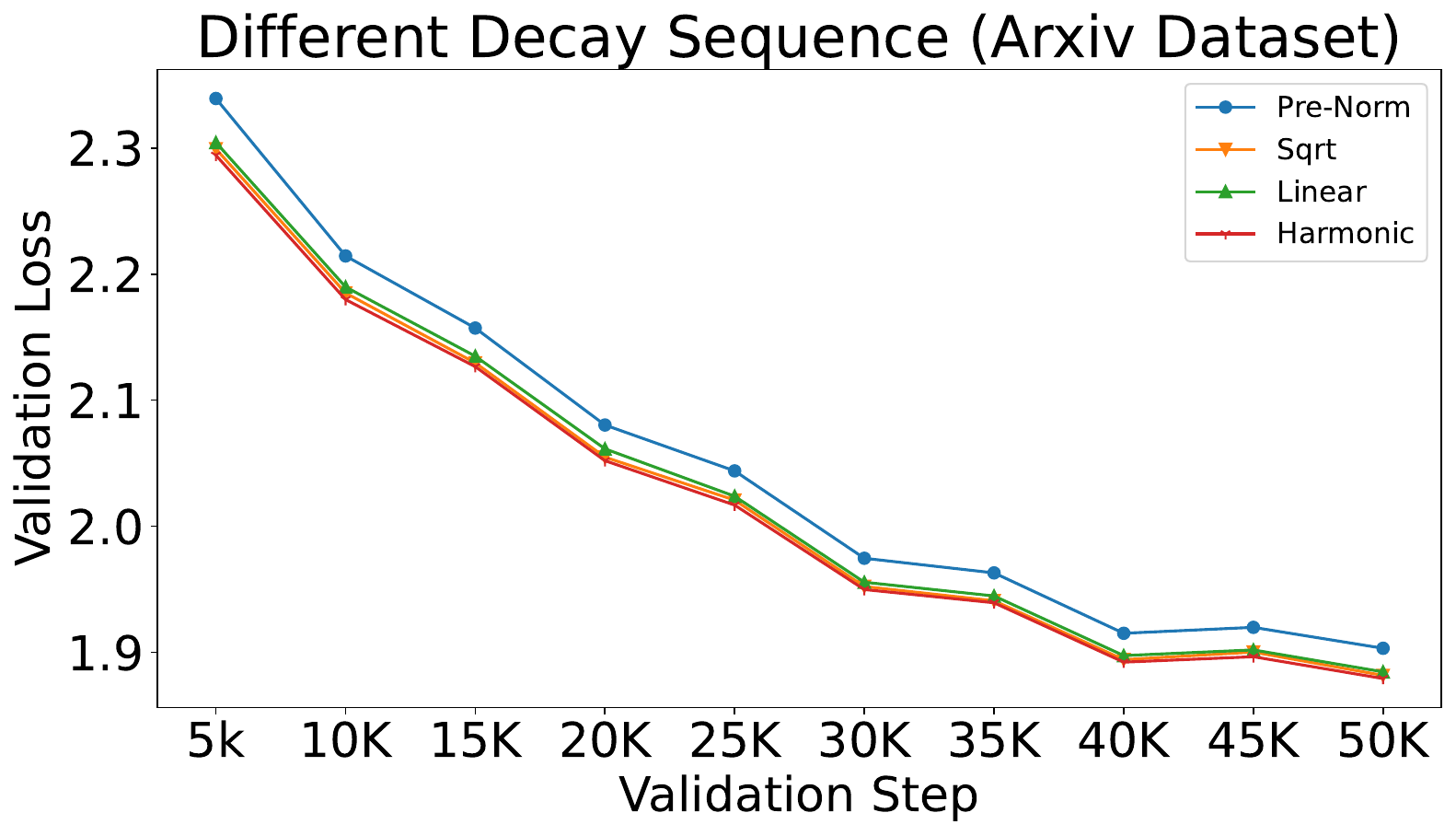}
\hspace{0in}
\includegraphics[width=0.45\textwidth]{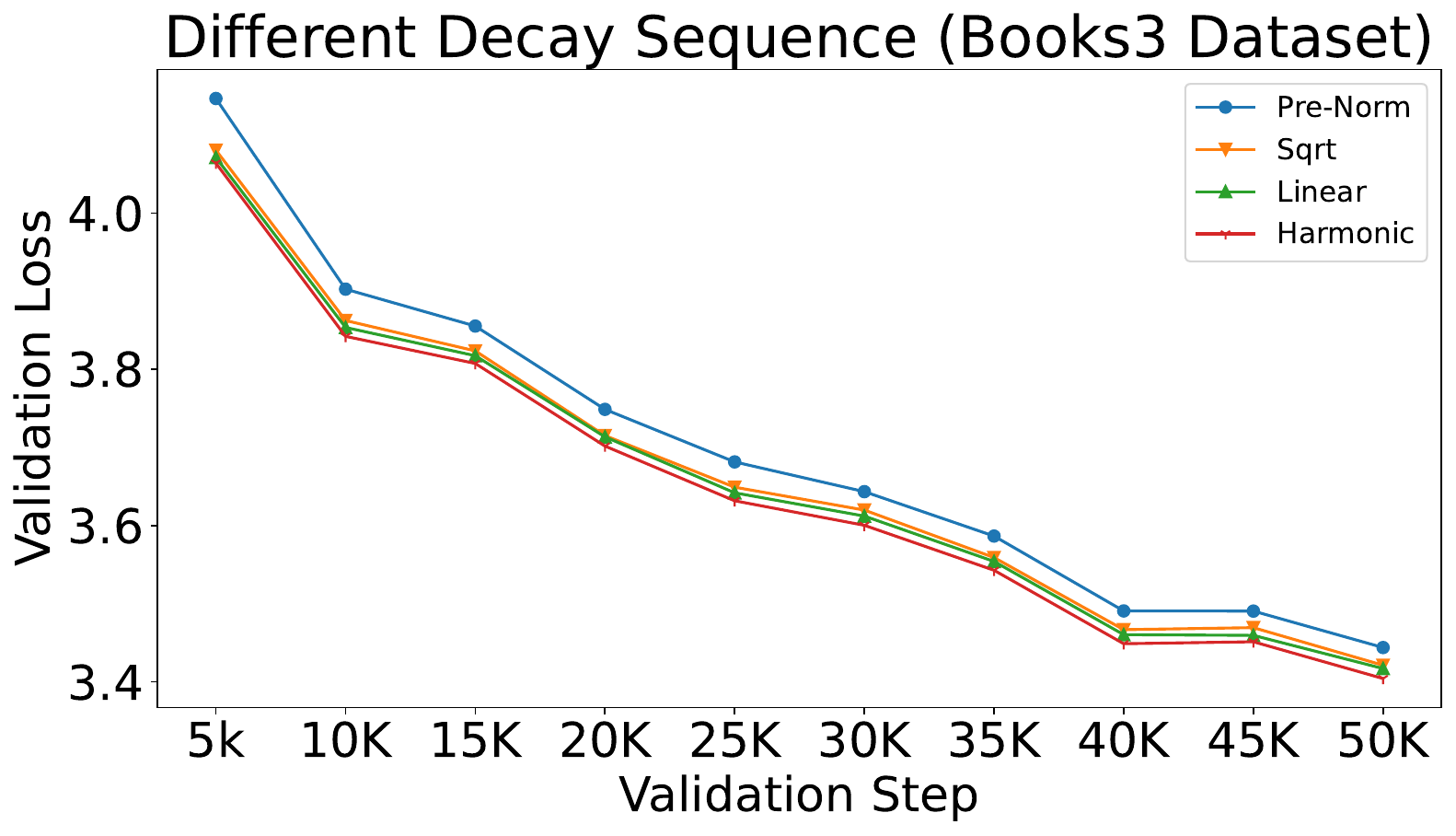}
\caption{
The performance of different decay methods on the Arxiv and Books3 dataset, with model parameter 125M, training lengths of 512. \textbf{Sqrt}: $\frac{\alpha}{k^{0.5}}$, \textbf{Linear}: $\frac{\alpha(T-k)}{T}$, and \textbf{Harnomic}: $\frac{\alpha}{k}$, where $k$ is the current layer index and $T$ is the total layer number. 
}
% \vspace{-5pt}
\label{fig: decay method}
\end{figure*}

\begin{figure*}[!htbp]
%\begin{figure*}
\centering
\includegraphics[width=0.43\textwidth]{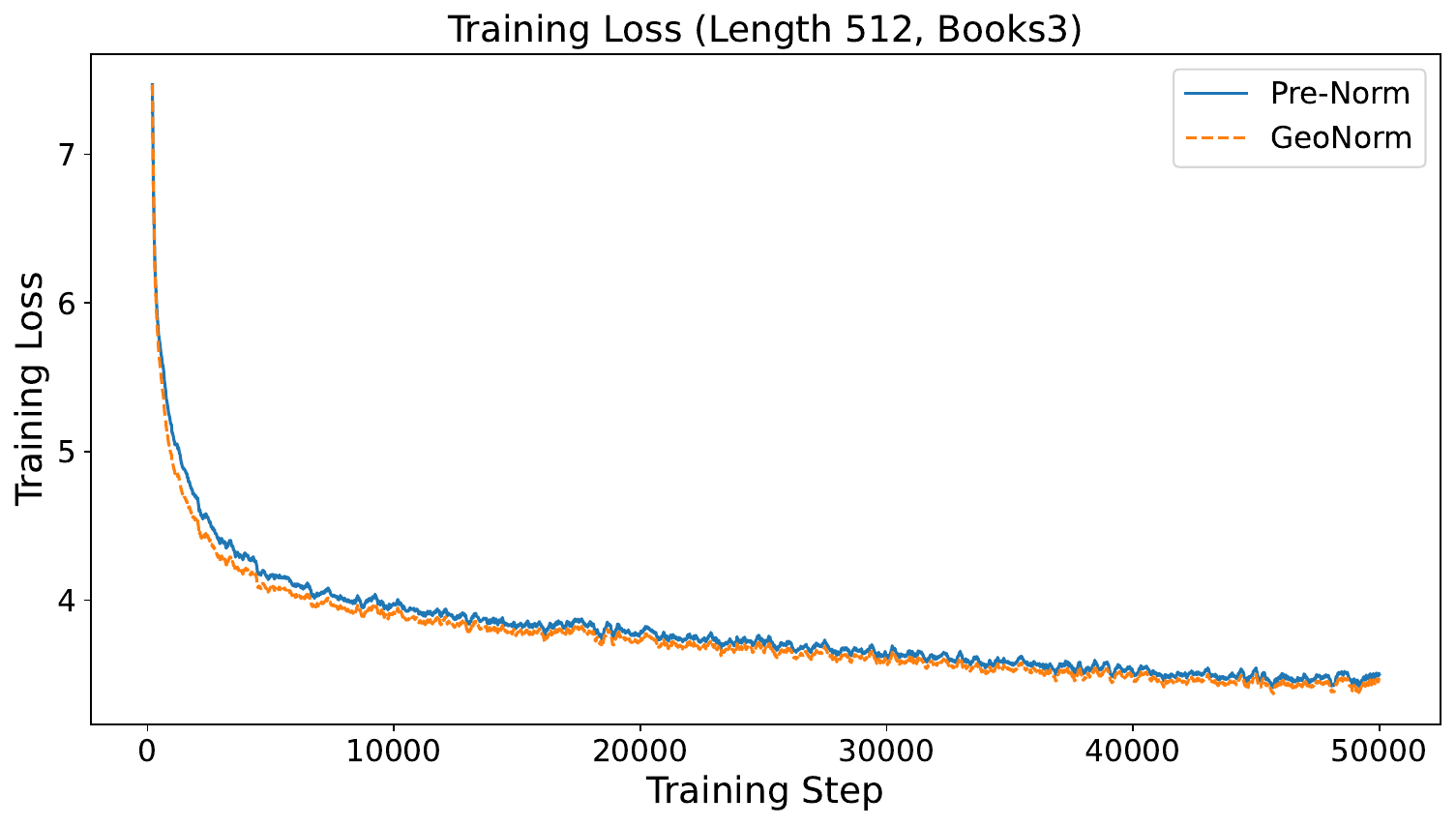}
\hspace{0in}
\includegraphics[width=0.45\textwidth]{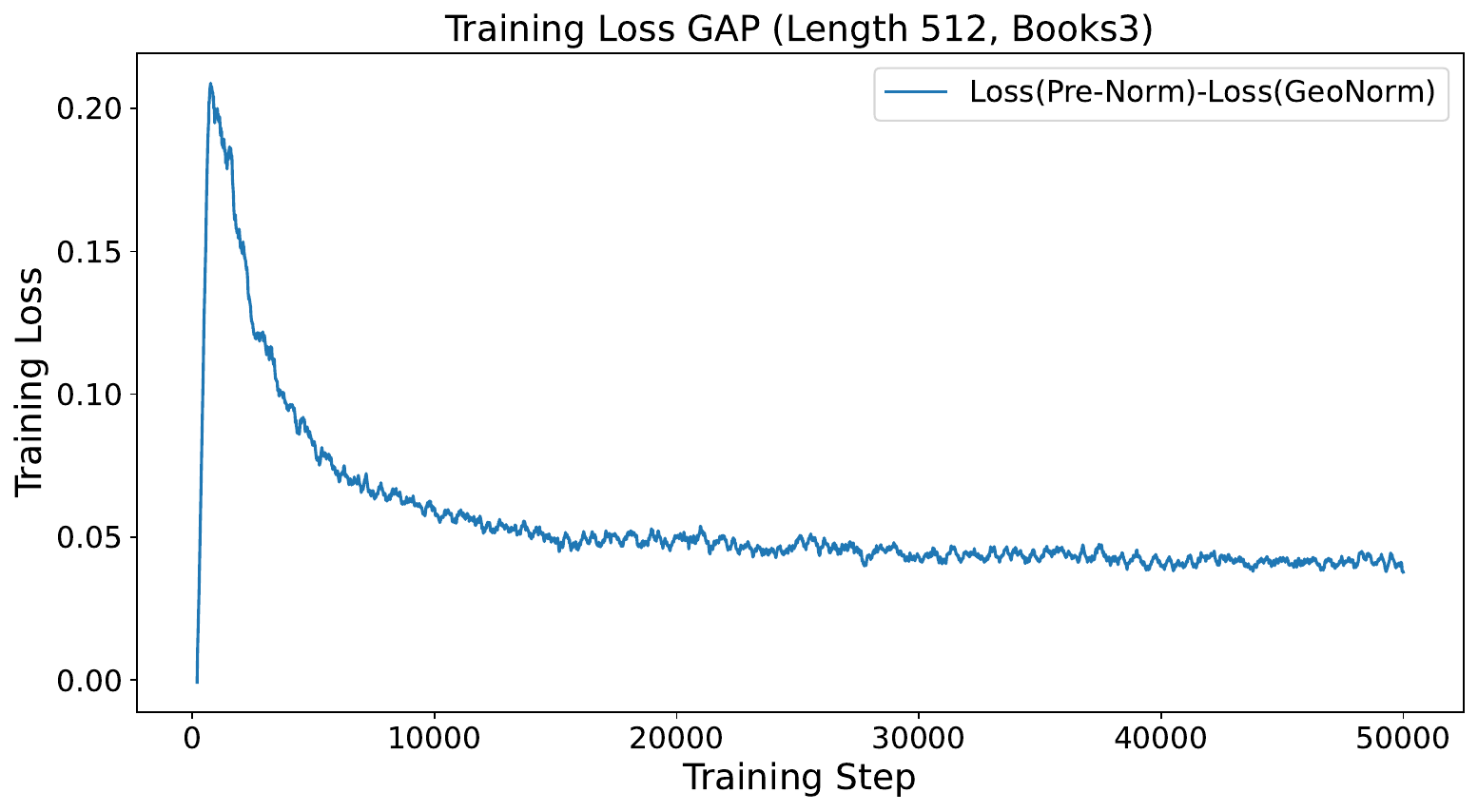}
\caption{
The training dynamic with loss metrics, with model parameter of 125M, training lengths of 512 and the Books3 dataset.
}
% \vspace{-5pt}
\label{fig: train loss}
\end{figure*}

\paragraph{The \methodShort with Hormonic Decay achieves better performance.} As shown in Figure \ref{fig: decay method}, the proposed \methodShort method demonstrates significantly enhanced performance when utilizing the harmonic decay scheduling mechanism. Across multiple benchmark datasets, this configuration consistently yields state-of-the-art results. 
Specifically, on the Arxiv dataset, \methodShort with harmonic decay achieves a loss of 1.8792, establishing a new performance benchmark. This represents a substantial improvement over alternative configurations, with the second-best performing method attaining only 1.8817. The performance improvement is even more pronounced on the larger and more complex Books3 dataset, where \methodShort with harmonic decay records a loss of 3.4048, again outperforming all competing approaches.
These empirical results collectively demonstrate that integrating harmonic decay scheduling with the \methodShort imprroves the performance.

\paragraph{The Hormonic Decay achieves the best performance across the validation steps.} On the Arxiv dataset, with the validation step 5000, the \methodShort achieves 2.2943 loss, which is the best performance, compared to Pre-Norm, DeepNorm and SandwichNorm. With the validation step increased to 50000, the \methodShort still achieves the lowest loss. Similarly, the \methodShort achieves the best performance on Books3 dataset from the validation step 5000 to 50000. Therefore, the \methodShort achieves the best performance across the validation steps.

\subsection{The Analysis of Training Dynamics}

\paragraph{The \methodShort achieves better performance across the training steps.} As shown in Figure~\ref{fig: train loss}, the \methodShort method establishes a decisive performance advantage early in the training phase, as evidenced by its significantly lower loss metrics. This initial lead is not merely temporary; \methodShort demonstrates remarkable stability by maintaining its optimal performance throughout the entire training cycle without degradation. In direct comparison, the Pre-Norm baseline is consistently outperformed, confirming that \methodShort's superiority is a persistent and defining characteristic of the training process from initialization to convergence.

\paragraph{The \methodShort has faster loss reduction.} The training dynamics reveal a significant loss gap with a peak value of 0.2. Crucially, this gap does not persist due to instability but consistently narrows as training progresses, converging to a much smaller final value of approximately 0.05 by the end of the run. This relatively large divergence indicates that the \methodShort method successfully reduces its loss at a markedly faster rate during the training phase compared to the baseline, suggesting more efficient optimization.

\section*{Broader Impact Statement}
This work focuses on improving the Transformer architecture, which may be helpful for better performance of the Transformer and further understanding of the Transformer.  And it should be noted that this work should not be abused.

% \section{Discussion}

% \paragraph{The Transformer process could be regarded as optimizing along the layer dimension.} The feature during the transformer could be regarded as the optimization, while the FFN and Attention are the gradient estimator. 

% \paragraph{ Bring better manifold optimization for GeoNorm.}  In the future, we could bring better mainfold optimization algorithms, such as Adam and so on, to the transformer layer-level feature update.Tabl

\section{Conclusion}
This work presents a unified view of Pre-Norm and Post-Norm through the lens of manifold optimization. Building on this formulation, we introduce \methodShort, a novel approach that leverages geodesic optimization to improve model performance. We conduct a comprehensive evaluation of \methodShort, testing it across multiple datasets, varying sequence lengths, different model sizes, and a range of downstream tasks. By advancing the theoretical understanding of normalization mechanisms, this research provides a foundation for subsequent innovations.

% \newpage

\nocite{*}

\bibliography{example_paper}

@article{bahdanau2014neural,
  title={Neural machine translation by jointly learning to align and translate},
  author={Bahdanau, Dzmitry and Cho, Kyunghyun and Bengio, Yoshua},
  journal={arXiv preprint arXiv:1409.0473},
  year={2014}
}

@article{xu2025seqpo,
  title={SeqPO-SiMT: Sequential Policy Optimization for Simultaneous Machine Translation},
  author={Xu, Ting and Huang, Zhichao and Sun, Jiankai and Cheng, Shanbo and Lam, Wai},
  journal={arXiv preprint arXiv:2505.20622},
  year={2025}
}

@inproceedings{zhang2020pegasus,
  title={Pegasus: Pre-training with extracted gap-sentences for abstractive summarization},
  author={Zhang, Jingqing and Zhao, Yao and Saleh, Mohammad and Liu, Peter},
  booktitle={International conference on machine learning},
  pages={11328--11339},
  year={2020},
  organization={PMLR}
}

@article{touvron2023llama,
  title={Llama: Open and efficient foundation language models},
  author={Touvron, Hugo and Lavril, Thibaut and Izacard, Gautier and Martinet, Xavier and Lachaux, Marie-Anne and Lacroix, Timoth{\'e}e and Rozi{\`e}re, Baptiste and Goyal, Naman and Hambro, Eric and Azhar, Faisal and others},
  journal={arXiv preprint arXiv:2302.13971},
  year={2023}
}

@article{fedus2022switch,
  title={Switch transformers: Scaling to trillion parameter models with simple and efficient sparsity},
  author={Fedus, William and Zoph, Barret and Shazeer, Noam},
  journal={Journal of Machine Learning Research},
  volume={23},
  number={120},
  pages={1--39},
  year={2022}
}

@inproceedings{
puigcerver2023sparse,
title={From Sparse to Soft Mixtures of Experts},
author={Joan Puigcerver and Carlos Riquelme Ruiz and Basil Mustafa and Neil Houlsby},
booktitle={The Twelfth International Conference on Learning Representations},
year={2024},
url={https://openreview.net/forum?id=jxpsAj7ltE}
}

@article{jiang2024mixtral,
  title={Mixtral of experts},
  author={Jiang, Albert Q and Sablayrolles, Alexandre and Roux, Antoine and Mensch, Arthur and Savary, Blanche and Bamford, Chris and Chaplot, Devendra Singh and Casas, Diego de las and Hanna, Emma Bou and Bressand, Florian and others},
  journal={arXiv preprint arXiv:2401.04088},
  year={2024}
}

@misc{meta2025llama,
  title={The llama 4 herd: The beginning of a new era of natively multimodal ai innovation},
  author={Meta, AI},
  howpublished={\url{https://ai. meta. com/blog/llama-4-multimodal-intelligence/}},
    note = {Accessed: 4-7-2025},
}

@article{liu2024deepseek,
  title={Deepseek-v3 technical report},
  author={Liu, Aixin and Feng, Bei and Xue, Bing and Wang, Bingxuan and Wu, Bochao and Lu, Chengda and Zhao, Chenggang and Deng, Chengqi and Zhang, Chenyu and Ruan, Chong and others},
  journal={arXiv preprint arXiv:2412.19437},
  year={2024}
}

@article{team2025kimi,
  title={Kimi k2: Open agentic intelligence},
  author={Team, Kimi and Bai, Yifan and Bao, Yiping and Chen, Guanduo and Chen, Jiahao and Chen, Ningxin and Chen, Ruijue and Chen, Yanru and Chen, Yuankun and Chen, Yutian and others},
  journal={arXiv preprint arXiv:2507.20534},
  year={2025}
}

@article{riquelme2021scaling,
  title={Scaling vision with sparse mixture of experts},
  author={Riquelme, Carlos and Puigcerver, Joan and Mustafa, Basil and Neumann, Maxim and Jenatton, Rodolphe and Susano Pinto, Andr{\'e} and Keysers, Daniel and Houlsby, Neil},
  journal={Advances in Neural Information Processing Systems},
  volume={34},
  pages={8583--8595},
  year={2021}
}

@inproceedings{lin2023video,
    title = "Video-{LL}a{VA}: Learning United Visual Representation by Alignment Before Projection",
    author = "Lin, Bin  and
      Ye, Yang  and
      Zhu, Bin  and
      Cui, Jiaxi  and
      Ning, Munan  and
      Jin, Peng  and
      Yuan, Li",
    editor = "Al-Onaizan, Yaser  and
      Bansal, Mohit  and
      Chen, Yun-Nung",
    booktitle = "Proceedings of the 2024 Conference on Empirical Methods in Natural Language Processing",
    month = nov,
    year = "2024",
    address = "Miami, Florida, USA",
    publisher = "Association for Computational Linguistics",
    url = "https://aclanthology.org/2024.emnlp-main.342/",
    doi = "10.18653/v1/2024.emnlp-main.342",
    pages = "5971--5984",
}

@inproceedings{he2016deep,
  title={Deep residual learning for image recognition},
  author={He, Kaiming and Zhang, Xiangyu and Ren, Shaoqing and Sun, Jian},
  booktitle={Proceedings of the IEEE conference on computer vision and pattern recognition},
  pages={770--778},
  year={2016}
}

@inproceedings{nguyen2019transformers,
  title={Transformers without Tears: Improving the Normalization of Self-Attention},
  author={Nguyen, Toan Q and Salazar, Julian},
  booktitle={Proceedings of the 16th International Conference on Spoken Language Translation},
  year={2019}
}

@article{yang2025qwen3,
  title={Qwen3 technical report},
  author={Yang, An and Li, Anfeng and Yang, Baosong and Zhang, Beichen and Hui, Binyuan and Zheng, Bo and Yu, Bowen and Gao, Chang and Huang, Chengen and Lv, Chenxu and others},
  journal={arXiv preprint arXiv:2505.09388},
  year={2025}
}

@article{popel2018training,
  title={Training tips for the transformer model},
  author={Popel, Martin and Bojar, Ond{\v{r}}ej},
  journal={arXiv preprint arXiv:1804.00247},
  year={2018}
}

@inproceedings{shazeer2018adafactor,
  title={Adafactor: Adaptive learning rates with sublinear memory cost},
  author={Shazeer, Noam and Stern, Mitchell},
  booktitle={International Conference on Machine Learning},
  pages={4596--4604},
  year={2018},
  organization={PMLR}
}

@article{wang2024deepnet,
  title={Deepnet: Scaling transformers to 1,000 layers},
  author={Wang, Hongyu and Ma, Shuming and Dong, Li and Huang, Shaohan and Zhang, Dongdong and Wei, Furu},
  journal={IEEE Transactions on Pattern Analysis and Machine Intelligence},
  volume={46},
  number={10},
  pages={6761--6774},
  year={2024},
  publisher={IEEE}
}

@article{ding2021cogview,
  title={Cogview: Mastering text-to-image generation via transformers},
  author={Ding, Ming and Yang, Zhuoyi and Hong, Wenyi and Zheng, Wendi and Zhou, Chang and Yin, Da and Lin, Junyang and Zou, Xu and Shao, Zhou and Yang, Hongxia and others},
  journal={Advances in neural information processing systems},
  volume={34},
  pages={19822--19835},
  year={2021}
}

@article{yin2025pangu,
  title={Pangu ultra: Pushing the limits of dense large language models on ascend npus},
  author={Yin, Yichun and Huang, Wenyong and Song, Kaikai and Tang, Yehui and Wu, Xueyu and Guo, Wei and Guo, Peng and Wang, Yaoyuan and Meng, Xiaojun and Wang, Yasheng and others},
  journal={arXiv preprint arXiv:2504.07866},
  year={2025}
}

@misc{lozhkov2024fineweb-edu,
    author       = { Lozhkov, Anton and Ben Allal, Loubna and von Werra, Leandro and Wolf, Thomas },  
    title        = { FineWeb-Edu: the Finest Collection of Educational Content }, 
    year         = 2024,  
    url          = { https://huggingface.co/datasets/HuggingFaceFW/fineweb-edu },  
    doi          = { 10.57967/hf/2497 },
    publisher    = { Hugging Face }
}

@article{clark2018think,
  title={Think you have solved question answering? try arc, the ai2 reasoning challenge},
  author={Clark, Peter and Cowhey, Isaac and Etzioni, Oren and Khot, Tushar and Sabharwal, Ashish and Schoenick, Carissa and Tafjord, Oyvind},
  journal={arXiv preprint arXiv:1803.05457},
  year={2018}
}

@inproceedings{zellers2019hellaswag,
    title = "{H}ella{S}wag: Can a Machine Really Finish Your Sentence?",
    author = "Zellers, Rowan  and
      Holtzman, Ari  and
      Bisk, Yonatan  and
      Farhadi, Ali  and
      Choi, Yejin",
    editor = "Korhonen, Anna  and
      Traum, David  and
      M{\`a}rquez, Llu{\'i}s",
    booktitle = "Proceedings of the 57th Annual Meeting of the Association for Computational Linguistics",
    month = jul,
    year = "2019",
    address = "Florence, Italy",
    publisher = "Association for Computational Linguistics",
    url = "https://aclanthology.org/P19-1472/",
    doi = "10.18653/v1/P19-1472",
    pages = "4791--4800",
}

@inproceedings{bisk2020piqa,
  title={Piqa: Reasoning about physical commonsense in natural language},
  author={Bisk, Yonatan and Zellers, Rowan and Gao, Jianfeng and Choi, Yejin and others},
  booktitle={Proceedings of the AAAI conference on artificial intelligence},
  volume={34},
  number={05},
  pages={7432--7439},
  year={2020}
}

@inproceedings{welbl2017crowdsourcing,
    title = "Crowdsourcing Multiple Choice Science Questions",
    author = "Welbl, Johannes  and
      Liu, Nelson F.  and
      Gardner, Matt",
    editor = "Derczynski, Leon  and
      Xu, Wei  and
      Ritter, Alan  and
      Baldwin, Tim",
    booktitle = "Proceedings of the 3rd Workshop on Noisy User-generated Text",
    month = sep,
    year = "2017",
    address = "Copenhagen, Denmark",
    publisher = "Association for Computational Linguistics",
    url = "https://aclanthology.org/W17-4413/",
    doi = "10.18653/v1/W17-4413",
    pages = "94--106",
}

@article{sakaguchi2021winogrande,
  title={Winogrande: An adversarial winograd schema challenge at scale},
  author={Sakaguchi, Keisuke and Bras, Ronan Le and Bhagavatula, Chandra and Choi, Yejin},
  journal={Communications of the ACM},
  volume={64},
  number={9},
  pages={99--106},
  year={2021},
  publisher={ACM New York, NY, USA}
}

@misc{eval-harness,
  author       = {Gao, Leo and Tow, Jonathan and Abbasi, Baber and Biderman, Stella and Black, Sid and DiPofi, Anthony and Foster, Charles and Golding, Laurence and Hsu, Jeffrey and Le Noac'h, Alain and Li, Haonan and McDonell, Kyle and Muennighoff, Niklas and Ociepa, Chris and Phang, Jason and Reynolds, Laria and Schoelkopf, Hailey and Skowron, Aviya and Sutawika, Lintang and Tang, Eric and Thite, Anish and Wang, Ben and Wang, Kevin and Zou, Andy},
  title        = {The Language Model Evaluation Harness},
  month        = 07,
  year         = 2024,
  publisher    = {Zenodo},
  version      = {v0.4.3},
  doi          = {10.5281/zenodo.12608602},
  url          = {https://zenodo.org/records/12608602}
}

@article{liu2020very,
  title={Very deep transformers for neural machine translation},
  author={Liu, Xiaodong and Duh, Kevin and Liu, Liyuan and Gao, Jianfeng},
  journal={arXiv preprint arXiv:2008.07772},
  year={2020}
}

@article{brown2020language,
  title={Language models are few-shot learners},
  author={Brown, Tom and Mann, Benjamin and Ryder, Nick and Subbiah, Melanie and Kaplan, Jared D and Dhariwal, Prafulla and Neelakantan, Arvind and Shyam, Pranav and Sastry, Girish and Askell, Amanda and others},
  journal={Advances in neural information processing systems},
  volume={33},
  pages={1877--1901},
  year={2020}
}

@article{ouyang2022training,
  title={Training language models to follow instructions with human feedback},
  author={Ouyang, Long and Wu, Jeffrey and Jiang, Xu and Almeida, Diogo and Wainwright, Carroll and Mishkin, Pamela and Zhang, Chong and Agarwal, Sandhini and Slama, Katarina and Ray, Alex and others},
  journal={Advances in neural information processing systems},
  volume={35},
  pages={27730--27744},
  year={2022}
}

@article{bottou1998online,
  title={Online algorithms and stochastic approximations},
  author={Bottou, L{\'e}on},
  journal={Online learning in neural networks},
  year={1998},
  publisher={Cambridge University Press}
}

@article{rumelhart1986learning,
  title={Learning representations by back-propagating errors},
  author={Rumelhart, David E and Hinton, Geoffrey E and Williams, Ronald J},
  journal={nature},
  volume={323},
  number={6088},
  pages={533--536},
  year={1986},
  publisher={Nature Publishing Group UK London}
}

@article{duchi2011adaptive,
  title={Adaptive subgradient methods for online learning and stochastic optimization.},
  author={Duchi, John and Hazan, Elad and Singer, Yoram},
  journal={Journal of machine learning research},
  volume={12},
  number={7},
  year={2011}
}

@article{graves2013generating,
  title={Generating sequences with recurrent neural networks},
  author={Graves, Alex},
  journal={arXiv preprint arXiv:1308.0850},
  year={2013}
}

@misc{kingma2017adammethodstochasticoptimization,
      title={Adam: A Method for Stochastic Optimization}, 
      author={Diederik P. Kingma and Jimmy Ba},
      year={2017},
      eprint={1412.6980},
      archivePrefix={arXiv},
      primaryClass={cs.LG},
      url={https://arxiv.org/abs/1412.6980}, 
}

@article{hu2020brief,
  title={A brief introduction to manifold optimization},
  author={Hu, Jiang and Liu, Xin and Wen, Zai-Wen and Yuan, Ya-Xiang},
  journal={Journal of the Operations Research Society of China},
  volume={8},
  number={2},
  pages={199--248},
  year={2020},
  publisher={Springer}
}

@article{debye1909naherungsformeln,
  title={N{\"a}herungsformeln f{\"u}r die Zylinderfunktionen f{\"u}r gro{\ss}e Werte des Arguments und unbeschr{\"a}nkt ver{\"a}nderliche Werte des Index},
  author={Debye, Peter},
  journal={Mathematische Annalen},
  volume={67},
  number={4},
  pages={535--558},
  year={1909},
  publisher={Springer}
}

@article{da2025thorough,
  title={A thorough study of Riemannian Newton's Method},
  author={da Silva, Caio O and Aoto, Yuri A and Costa, Felipe FGS and da Silva, M{\'a}rcio F},
  journal={arXiv preprint arXiv:2506.09297},
  year={2025}
}

@article{hestenes1952methods,
  title={Methods of conjugate gradients for solving linear systems},
  author={Hestenes, Magnus R and Stiefel, Eduard and others},
  journal={Journal of research of the National Bureau of Standards},
  volume={49},
  number={6},
  pages={409--436},
  year={1952}
}

@inproceedings{zhang2016first,
  title={First-order methods for geodesically convex optimization},
  author={Zhang, Hongyi and Sra, Suvrit},
  booktitle={Conference on learning theory},
  pages={1617--1638},
  year={2016},
  organization={PMLR}
}

@article{ainslie2023gqa,
  title={Gqa: Training generalized multi-query transformer models from multi-head checkpoints},
  author={Ainslie, Joshua and Lee-Thorp, James and De Jong, Michiel and Zemlyanskiy, Yury and Lebr{\'o}n, Federico and Sanghai, Sumit},
  journal={arXiv preprint arXiv:2305.13245},
  year={2023}
}

@article{mihaylov2018can,
  title={Can a suit of armor conduct electricity? a new dataset for open book question answering},
  author={Mihaylov, Todor and Clark, Peter and Khot, Tushar and Sabharwal, Ashish},
  journal={arXiv preprint arXiv:1809.02789},
  year={2018}
}

@article{sap2019socialiqa,
  title={Socialiqa: Commonsense reasoning about social interactions},
  author={Sap, Maarten and Rashkin, Hannah and Chen, Derek and LeBras, Ronan and Choi, Yejin},
  journal={arXiv preprint arXiv:1904.09728},
  year={2019}
}

@article{clark2019boolq,
  title={Boolq: Exploring the surprising difficulty of natural yes/no questions},
  author={Clark, Christopher and Lee, Kenton and Chang, Ming-Wei and Kwiatkowski, Tom and Collins, Michael and Toutanova, Kristina},
  journal={arXiv preprint arXiv:1905.10044},
  year={2019}
}

@article{chai2020highway,
  title={Highway transformer: Self-gating enhanced self-attentive networks},
  author={Chai, Yekun and Jin, Shuo and Hou, Xinwen},
  journal={arXiv preprint arXiv:2004.08178},
  year={2020}
}

@article{fang2023cross,
  title={Cross-layer retrospective retrieving via layer attention},
  author={Fang, Yanwen and Cai, Yuxi and Chen, Jintai and Zhao, Jingyu and Tian, Guangjian and Li, Guodong},
  journal={arXiv preprint arXiv:2302.03985},
  year={2023}
}

@article{mak2025residual,
  title={Residual Matrix Transformers: Scaling the Size of the Residual Stream},
  author={Mak, Brian and Flanigan, Jeffrey},
  journal={arXiv preprint arXiv:2506.22696},
  year={2025}
}

@inproceedings{chollet2017xception,
  title={Xception: Deep learning with depthwise separable convolutions},
  author={Chollet, Fran{\c{c}}ois},
  booktitle={Proceedings of the IEEE conference on computer vision and pattern recognition},
  pages={1251--1258},
  year={2017}
}

@article{sun2025survey,
  title={A survey of reasoning with foundation models: Concepts, methodologies, and outlook},
  author={Sun, Jiankai and Zheng, Chuanyang and Xie, Enze and Liu, Zhengying and Chu, Ruihang and Qiu, Jianing and Xu, Jiaqi and Ding, Mingyu and Li, Hongyang and Geng, Mengzhe and others},
  journal={ACM Computing Surveys},
  volume={57},
  number={11},
  pages={1--43},
  year={2025},
  publisher={ACM New York, NY}
}

@article{cobbe2021training,
  title={Training verifiers to solve math word problems},
  author={Cobbe, Karl and Kosaraju, Vineet and Bavarian, Mohammad and Chen, Mark and Jun, Heewoo and Kaiser, Lukasz and Plappert, Matthias and Tworek, Jerry and Hilton, Jacob and Nakano, Reiichiro and others},
  journal={arXiv preprint arXiv:2110.14168},
  year={2021}
}

@article{li2024mix,
  title={Mix-ln: Unleashing the power of deeper layers by combining pre-ln and post-ln},
  author={Li, Pengxiang and Yin, Lu and Liu, Shiwei},
  journal={arXiv preprint arXiv:2412.13795},
  year={2024}
}

@article{zhuo2025hybridnorm,
  title={HybridNorm: Towards Stable and Efficient Transformer Training via Hybrid Normalization},
  author={Zhuo, Zhijian and Zeng, Yutao and Wang, Ya and Zhang, Sijun and Yang, Jian and Li, Xiaoqing and Zhou, Xun and Ma, Jinwen},
  journal={arXiv preprint arXiv:2503.04598},
  year={2025}
}

@article{menary2024transformer,
  title={Transformer Normalisation Layers and the Independence of Semantic Subspaces},
  author={Menary, Stephen and Kaski, Samuel and Freitas, Andre},
  journal={arXiv preprint arXiv:2406.17837},
  year={2024}
}

@article{rybakov2024methods,
  title={Methods of improving llm training stability},
  author={Rybakov, Oleg and Chrzanowski, Mike and Dykas, Peter and Xue, Jinze and Lanir, Ben},
  journal={arXiv preprint arXiv:2410.16682},
  year={2024}
}

@article{shleifer2021normformer,
  title={Normformer: Improved transformer pretraining with extra normalization},
  author={Shleifer, Sam and Weston, Jason and Ott, Myle},
  journal={arXiv preprint arXiv:2110.09456},
  year={2021}
}

@article{zhang2019improving,
  title={Improving deep transformer with depth-scaled initialization and merged attention},
  author={Zhang, Biao and Titov, Ivan and Sennrich, Rico},
  journal={arXiv preprint arXiv:1908.11365},
  year={2019}
}

@inproceedings{huang2020improving,
  title={Improving transformer optimization through better initialization},
  author={Huang, Xiao Shi and Perez, Felipe and Ba, Jimmy and Volkovs, Maksims},
  booktitle={International Conference on Machine Learning},
  pages={4475--4483},
  year={2020},
  organization={PMLR}
}

@article{ba2016layer,
  title={Layer Normalization},
  author={Ba, Jimmy Lei and Kiros, Jamie Ryan and Hinton, Geoffrey E},
  year={2016}
}

@article{zhang2019root,
  title={Root mean square layer normalization},
  author={Zhang, Biao and Sennrich, Rico},
  journal={Advances in neural information processing systems},
  volume={32},
  year={2019}
}

@inproceedings{dehghani2023scaling,
  title={Scaling vision transformers to 22 billion parameters},
  author={Dehghani, Mostafa and Djolonga, Josip and Mustafa, Basil and Padlewski, Piotr and Heek, Jonathan and Gilmer, Justin and Steiner, Andreas Peter and Caron, Mathilde and Geirhos, Robert and Alabdulmohsin, Ibrahim and others},
  booktitle={International conference on machine learning},
  pages={7480--7512},
  year={2023},
  organization={PMLR}
}

@inproceedings{henry2020query,
  title={Query-key normalization for transformers},
  author={Henry, Alex and Dachapally, Prudhvi Raj and Pawar, Shubham Shantaram and Chen, Yuxuan},
  booktitle={Findings of the Association for Computational Linguistics: EMNLP 2020},
  pages={4246--4253},
  year={2020}
}

@article{klein2017opennmt,
  title={Opennmt: Open-source toolkit for neural machine translation},
  author={Klein, Guillaume and Kim, Yoon and Deng, Yuntian and Senellart, Jean and Rush, Alexander M},
  journal={arXiv preprint arXiv:1701.02810},
  year={2017}
}

@article{liu2020understanding,
  title={Understanding the difficulty of training transformers},
  author={Liu, Liyuan and Liu, Xiaodong and Gao, Jianfeng and Chen, Weizhu and Han, Jiawei},
  journal={arXiv preprint arXiv:2004.08249},
  year={2020}
}

@article{muennighoff2024olmoe,
  title={Olmoe: Open mixture-of-experts language models},
  author={Muennighoff, Niklas and Soldaini, Luca and Groeneveld, Dirk and Lo, Kyle and Morrison, Jacob and Min, Sewon and Shi, Weijia and Walsh, Pete and Tafjord, Oyvind and Lambert, Nathan and others},
  journal={arXiv preprint arXiv:2409.02060},
  year={2024}
}

@article{paszke2019pytorch,
  title={Pytorch: An imperative style, high-performance deep learning library},
  author={Paszke, Adam and Gross, Sam and Massa, Francisco and Lerer, Adam and Bradbury, James and Chanan, Gregory and Killeen, Trevor and Lin, Zeming and Gimelshein, Natalia and Antiga, Luca and others},
  journal={Advances in neural information processing systems},
  volume={32},
  year={2019}
}

@inproceedings{takase2023b2t,
  title={B2t connection: Serving stability and performance in deep transformers},
  author={Takase, Sho and Kiyono, Shun and Kobayashi, Sosuke and Suzuki, Jun},
  booktitle={Findings of the Association for Computational Linguistics: ACL 2023},
  pages={3078--3095},
  year={2023}
}

@article{kedia2024transformers,
  title={Transformers get stable: An end-to-end signal propagation theory for language models},
  author={Kedia, Akhil and Zaidi, Mohd Abbas and Khyalia, Sushil and Jung, Jungho and Goka, Harshith and Lee, Haejun},
  journal={arXiv preprint arXiv:2403.09635},
  year={2024}
}

@article{team2024gemma,
  title={Gemma 2: Improving open language models at a practical size},
  author={Team, Gemma and Riviere, Morgane and Pathak, Shreya and Sessa, Pier Giuseppe and Hardin, Cassidy and Bhupatiraju, Surya and Hussenot, L{\'e}onard and Mesnard, Thomas and Shahriari, Bobak and Ram{\'e}, Alexandre and others},
  journal={arXiv preprint arXiv:2408.00118},
  year={2024}
}

@article{wang2019learning,
  title={Learning deep transformer models for machine translation},
  author={Wang, Qiang and Li, Bei and Xiao, Tong and Zhu, Jingbo and Li, Changliang and Wong, Derek F and Chao, Lidia S},
  journal={arXiv preprint arXiv:1906.01787},
  year={2019}
}

@article{bjorck2018understanding,
  title={Understanding batch normalization},
  author={Bjorck, Nils and Gomes, Carla P and Selman, Bart and Weinberger, Kilian Q},
  journal={Advances in neural information processing systems},
  volume={31},
  year={2018}
}

@article{loshchilov2024ngpt,
  title={ngpt: Normalized transformer with representation learning on the hypersphere},
  author={Loshchilov, Ilya and Hsieh, Cheng-Ping and Sun, Simeng and Ginsburg, Boris},
  journal={arXiv preprint arXiv:2410.01131},
  year={2024}
}

@article{ulyanov2016instance,
  title={Instance normalization: The missing ingredient for fast stylization},
  author={Ulyanov, Dmitry and Vedaldi, Andrea and Lempitsky, Victor},
  journal={arXiv preprint arXiv:1607.08022},
  year={2016}
}

@article{konstantinidis2023multi,
  title={Multi-manifold attention for vision transformers},
  author={Konstantinidis, Dimitrios and Papastratis, Ilias and Dimitropoulos, Kosmas and Daras, Petros},
  journal={IEEE Access},
  volume={11},
  pages={123433--123444},
  year={2023},
  publisher={IEEE}
}

@article{larsson2016fractalnet,
  title={Fractalnet: Ultra-deep neural networks without residuals},
  author={Larsson, Gustav and Maire, Michael and Shakhnarovich, Gregory},
  journal={arXiv preprint arXiv:1605.07648},
  year={2016}
}

@article{loshchilov2017decoupled,
  title={Decoupled weight decay regularization},
  author={Loshchilov, Ilya and Hutter, Frank},
  journal={arXiv preprint arXiv:1711.05101},
  year={2017}
}

@article{menghani2024laurel,
  title={LAUREL: Learned Augmented Residual Layer},
  author={Menghani, Gaurav and Kumar, Ravi and Kumar, Sanjiv},
  journal={arXiv preprint arXiv:2411.07501},
  year={2024}
}

@inproceedings{wu2018group,
  title={Group normalization},
  author={Wu, Yuxin and He, Kaiming},
  booktitle={Proceedings of the European conference on computer vision (ECCV)},
  pages={3--19},
  year={2018}
}

@article{kim2025peri,
  title={Peri-ln: Revisiting normalization layer in the transformer architecture},
  author={Kim, Jeonghoon and Lee, Byeongchan and Park, Cheonbok and Oh, Yeontaek and Kim, Beomjun and Yoo, Taehwan and Shin, Seongjin and Han, Dongyoon and Shin, Jinwoo and Yoo, Kang Min},
  journal={arXiv preprint arXiv:2502.02732},
  year={2025}
}

@article{pagliardini2024denseformer,
  title={Denseformer: Enhancing information flow in transformers via depth weighted averaging},
  author={Pagliardini, Matteo and Mohtashami, Amirkeivan and Fleuret, Francois and Jaggi, Martin},
  journal={Advances in neural information processing systems},
  volume={37},
  pages={136479--136508},
  year={2024}
}

@article{li2020shallow,
  title={Shallow-to-deep training for neural machine translation},
  author={Li, Bei and Wang, Ziyang and Liu, Hui and Jiang, Yufan and Du, Quan and Xiao, Tong and Wang, Huizhen and Zhu, Jingbo},
  journal={arXiv preprint arXiv:2010.03737},
  year={2020}
}

@article{shoeybi2019megatron,
  title={Megatron-lm: Training multi-billion parameter language models using model parallelism},
  author={Shoeybi, Mohammad and Patwary, Mostofa and Puri, Raul and LeGresley, Patrick and Casper, Jared and Catanzaro, Bryan},
  journal={arXiv preprint arXiv:1909.08053},
  year={2019}
}

@article{takase2023spike,
  title={Spike no more: Stabilizing the pre-training of large language models},
  author={Takase, Sho and Kiyono, Shun and Kobayashi, Sosuke and Suzuki, Jun},
  journal={arXiv preprint arXiv:2312.16903},
  year={2023}
}

@inproceedings{xiong2020layer,
  title={On layer normalization in the transformer architecture},
  author={Xiong, Ruibin and Yang, Yunchang and He, Di and Zheng, Kai and Zheng, Shuxin and Xing, Chen and Zhang, Huishuai and Lan, Yanyan and Wang, Liwei and Liu, Tieyan},
  booktitle={International conference on machine learning},
  pages={10524--10533},
  year={2020},
  organization={PMLR}
}

@article{bapna2018training,
  title={Training deeper neural machine translation models with transparent attention},
  author={Bapna, Ankur and Chen, Mia Xu and Firat, Orhan and Cao, Yuan and Wu, Yonghui},
  journal={arXiv preprint arXiv:1808.07561},
  year={2018}
}

@article{gromov2024unreasonable,
  title={The unreasonable ineffectiveness of the deeper layers},
  author={Gromov, Andrey and Tirumala, Kushal and Shapourian, Hassan and Glorioso, Paolo and Roberts, Daniel A},
  journal={arXiv preprint arXiv:2403.17887},
  year={2024}
}

@inproceedings{hu2023llm,
  title={Llm-adapters: An adapter family for parameter-efficient fine-tuning of large language models},
  author={Hu, Zhiqiang and Wang, Lei and Lan, Yihuai and Xu, Wanyu and Lim, Ee-Peng and Bing, Lidong and Xu, Xing and Poria, Soujanya and Lee, Roy},
  booktitle={Proceedings of the 2023 conference on empirical methods in natural language processing},
  pages={5254--5276},
  year={2023}
}

@article{lialin2023relora,
  title={Relora: High-rank training through low-rank updates},
  author={Lialin, Vladislav and Shivagunde, Namrata and Muckatira, Sherin and Rumshisky, Anna},
  journal={arXiv preprint arXiv:2307.05695},
  year={2023}
}

@article{lialin2023stack,
  title={Stack more layers differently: High-rank training through low-rank updates},
  author={Lialin, Vladislav and Muckatira, Sherin and Shivagunde, Namrata and Rumshisky, Anna},
  year={2023}
}

@inproceedings{men2025shortgpt,
  title={Shortgpt: Layers in large language models are more redundant than you expect},
  author={Men, Xin and Xu, Mingyu and Zhang, Qingyu and Yuan, Qianhao and Wang, Bingning and Lin, Hongyu and Lu, Yaojie and Han, Xianpei and Chen, Weipeng},
  booktitle={Findings of the Association for Computational Linguistics: ACL 2025},
  pages={20192--20204},
  year={2025}
}

@article{siddiqui2024deeper,
  title={A deeper look at depth pruning of llms},
  author={Siddiqui, Shoaib Ahmed and Dong, Xin and Heinrich, Greg and Breuel, Thomas and Kautz, Jan and Krueger, David and Molchanov, Pavlo},
  journal={arXiv preprint arXiv:2407.16286},
  year={2024}
}

@article{zhang2024adam,
  title={Adam-mini: Use fewer learning rates to gain more},
  author={Zhang, Yushun and Chen, Congliang and Li, Ziniu and Ding, Tian and Wu, Chenwei and Kingma, Diederik P and Ye, Yinyu and Luo, Zhi-Quan and Sun, Ruoyu},
  journal={arXiv preprint arXiv:2406.16793},
  year={2024}
}

@article{zhao2024galore,
  title={Galore: Memory-efficient llm training by gradient low-rank projection},
  author={Zhao, Jiawei and Zhang, Zhenyu and Chen, Beidi and Wang, Zhangyang and Anandkumar, Anima and Tian, Yuandong},
  journal={arXiv preprint arXiv:2403.03507},
  year={2024}
}

@article{jaiswal2024galore,
  title={From galore to welore: How low-rank weights non-uniformly emerge from low-rank gradients},
  author={Jaiswal, Ajay and Yin, Lu and Zhang, Zhenyu and Liu, Shiwei and Zhao, Jiawei and Tian, Yuandong and Wang, Zhangyang},
  journal={arXiv preprint arXiv:2407.11239},
  year={2024}
}

@article{liu2022more,
  title={More convnets in the 2020s: Scaling up kernels beyond 51x51 using sparsity},
  author={Liu, Shiwei and Chen, Tianlong and Chen, Xiaohan and Chen, Xuxi and Xiao, Qiao and Wu, Boqian and K{\"a}rkk{\"a}inen, Tommi and Pechenizkiy, Mykola and Mocanu, Decebal and Wang, Zhangyang},
  journal={arXiv preprint arXiv:2207.03620},
  year={2022}
}

@inproceedings{liu2022convnet,
  title={A convnet for the 2020s},
  author={Liu, Zhuang and Mao, Hanzi and Wu, Chao-Yuan and Feichtenhofer, Christoph and Darrell, Trevor and Xie, Saining},
  booktitle={Proceedings of the IEEE/CVF conference on computer vision and pattern recognition},
  pages={11976--11986},
  year={2022}
}
\bibliographystyle{icml2026}

%%%%%%%%%%%%%%%%%%%%%%%%%%%%%%%%%%%%%%%%%%%%%%%%%%%%%%%%%%%%%%%%%%%%%%%%%%%%%%%
%%%%%%%%%%%%%%%%%%%%%%%%%%%%%%%%%%%%%%%%%%%%%%%%%%%%%%%%%%%%%%%%%%%%%%%%%%%%%%%
% APPENDIX
%%%%%%%%%%%%%%%%%%%%%%%%%%%%%%%%%%%%%%%%%%%%%%%%%%%%%%%%%%%%%%%%%%%%%%%%%%%%%%%
%%%%%%%%%%%%%%%%%%%%%%%%%%%%%%%%%%%%%%%%%%%%%%%%%%%%%%%%%%%%%%%%%%%%%%%%%%%%%%%
\newpage
\appendix
\onecolumn

\section{Model Configuration}
\label{model configuration details} 
All experiments are conducted on 8 GPUs. The 125M and 350M model configuration is the following.
% The layer number is 12, the hidden dimension is 768 and the attention head is 12. The learning rate is 6e-4.

\begin{table}[htbp]
    \centering
    \setlength{\tabcolsep}{3pt}
    \label{model configuration}
    \caption{\textbf{Model Configurations.}}
    \resizebox{0.6\textwidth}{!}{
    \begin{tabular}{c c c c c}
    \toprule
    & & \textbf{125M} & & \textbf{350M} \\ \midrule
    Training sequence length & & $512$ & & $512$\\
    Batch size & & 32   & & 32  \\
    Number of iterations & & $50$k & & $50$k \\
    Dropout prob. & & $0.0$ & & $0.0$ \\
    Attention dropout prob. & & $0.0$ & & $0.0$ \\
    Attention head && 12 && 16  \\
    Feature dimension && 768 && 1024\\
    Layer number && 12 && 24 \\
    Optimizer & & Adam & & Adam\\
    Optimizer parameter betas & & [0.9, 0.95] && [0.9, 0.95] \\
    Learning rate & & $6\mathrm{e}-4$  & & $3\mathrm{e}-4$ \\
    Precision & & float16 & & float16 \\ 
    \bottomrule
    \end{tabular}
    }
    \label{tab:model_configs}
\end{table}

\section{Geodesic Optimization for Pre-Norm}
\label{appendix: prenorm}
Let $\bm{x}_k$ denote the token representation after the $k$-th Transformer layer.  
A Transformer layer with Pre-Norm can be expressed as:
\begin{equation}
    \begin{aligned}
        \tilde{\bm{x}}_k &= \bm{x}_k + \phi\bigl(\text{Norm}(\bm{x}_k)\bigr), \\
    \end{aligned}
    \label{eq:pre_norm_layer}
\end{equation}
where $\phi(\cdot)$ denotes either the attention module $\mathrm{Attn}(\cdot)$ or the feed-forward network $\mathrm{FFN}(\cdot)$, and $\mathrm{Norm}(\cdot)$ represents a normalization operator (typically RMSNorm).
% \begin{equation}
%     \bm{x}_k
% = \bm{x}
% + \phi_1(\mathrm{Norm}(\bm{x}))
% + \phi_2(\mathrm{Norm}(\bm{x}_1))
% + \cdots
% + \phi_k(\mathrm{Norm}(\bm{x}_{k-1})).
% \end{equation}
% After all layers, there is a normalization layer
% \begin{equation}
%     Norm(x)=Norm(\bm{x}/sqrt(k))
% = Norm(\bm{x}
% + \phi_1(\mathrm{Norm}(\bm{x}))
% + \phi_2(\mathrm{Norm}(\bm{x}_2))
% + \cdots
% + \phi_k(\mathrm{Norm}(\bm{x}_{k-1}))/sqrt(k)).
% \end{equation}

Let $\bm{x}_0 = \bm{x}$ be the initial input.
After $k$ layers of the Transformer model with a standard Pre-Norm, the representation can be written as
\begin{equation}
    \bm{x}_k = \bm{x} 
    + \phi_1\!\big(\mathrm{Norm}(\bm{x})\big) 
    + \phi_2\!\big(\mathrm{Norm}(\bm{x}_1)\big) 
    + \cdots 
    + \phi_k\!\big(\mathrm{Norm}(\bm{x}_{k-1})\big).
\end{equation}
This is obtained by recursively applying
\begin{equation}
\label{eq5}
    \bm{x}_{k} = \bm{x}_{k-1} + \phi_k\!\big(\mathrm{Norm}(\bm{x}_{k-1})\big),
\end{equation}
for each layer index $k$. Here, for simplicity, we use $\phi_k(\cdot)$ to denote the attention and feed-forward module at the $k$-th layer. 
At the output of the final layer, an additional normalization operation is applied:
\begin{equation}
    \text{FinalOutput} = \mathrm{Norm}\!\left(\bm{x}_{k}\right).
\end{equation}

We consider a widely used normalization method, namely \textbf{RMSNorm}, which is scale-invariant and normalizes vectors to have norm $\sqrt{D}$, where $D$ denotes the hidden dimension. Specifically,
\begin{equation}
    \mathrm{Norm}(\bm{v}) = \frac{\bm{v}}{\|\bm{v}\|}\sqrt{D}.
\end{equation}
As a consequence, for any scalar $c > 0$,
\begin{equation}
    \mathrm{Norm}\!\left(c\bm{v}\right) 
    = \frac{c\bm{v}}{\|c\bm{v}\|} \sqrt{D} 
    = \frac{\bm{v}}{\|\bm{v}\|} \sqrt{D}
    = \mathrm{Norm}(\bm{v}),
\end{equation}
showing that RMSNorm is invariant to positive rescaling.
Hence, we have
\begin{equation}
    \mathrm{Norm}\!\left(\frac{\bm{x}_k}{\sqrt{k+1}}\right) = \mathrm{Norm}(\bm{x}_k).
\end{equation}
This implies that, under RMSNorm, dividing the representation by scalars such as $\sqrt{k+1}$ has \textbf{no effect} on the final normalized output.

In standard Pre-Norm architectures, the norm $\|\bm{x}_k\|$ typically grows with depth $k$ due to the accumulation of residual updates in \eqref{eq5}. While one might attempt to control this growth by rescaling $\bm{x}_k$ (e.g., dividing by $\sqrt{k+1}$) prior to normalization, such scaling is entirely eliminated by RMSNorm.

% If $\mathrm{Norm}$ were LayerNorm (which subtracts the mean), then dividing by $\sqrt{k}$ would change the result.

% Assuming all parameters are initialized independently (e.g., with XXX initialization), these additive components are approximately orthogonal at initialization.
% Consequently, the norm of the accumulated representation satisfies
% \begin{equation*}
% \left\|
% \bm{x}
% + \sum_{i=1}^k \phi_i(\mathrm{Norm}(\bm{x}_{i-1}))
% \right\|
% \approx \sqrt{k}\,\|\bm{x}\|.
% \end{equation*}
% Since a normalization operation is applied before each layer, the following equivalence holds:
% \begin{equation*}
%     \mathrm{Norm}\!\left(
% \bm{x}
% + \sum_{i=1}^k \phi_i(\mathrm{Norm}(\bm{x}))
% \right)
% =
% \mathrm{Norm}\!\left(
% \frac{1}{\sqrt{k+1}}
% \left(
% \sqrt{k}\,\bm{x}
% + \phi(\mathrm{Norm}(\bm{x}))
% \right)
% \right).
% \end{equation*}

\paragraph{Complete Derivation of Alternative Scaled Update}

From the recursive update in \eqref{eq5}, we observe that
\begin{equation*}
    \frac{\bx_{k}}{\sqrt{k+1}} = \frac{\sqrt{k}}{\sqrt{k+1}} \frac{\bx_{k-1}}{\sqrt{k}} + \frac{1}{\sqrt{k+1}} \phi_k\left(\mathrm{Norm}\left(\frac{\bm{x}_{k-1}}{\sqrt{k}}\right)\right). 
\end{equation*}
Denote $\bm{x}_k^{\text{alt}}$ as $\frac{\bx_{k}}{\sqrt{k+1}}$, we obtain the alternative scaled update rule
\begin{equation}
\label{altscale}
    \bm{x}_k^{\text{alt}} = \frac{\sqrt{k}}{\sqrt{k+1}} \cdot \bm{x}_{k-1}^{\text{alt}}  + \frac{1}{\sqrt{k+1}} \cdot \phi_k\!\left(\mathrm{Norm}(\bm{x}_{k-1}^{\text{alt}} )\right), \quad k \ge 1
\end{equation}
with initialization $\bm{x}_0^{\text{alt}} = \bm{x}_0$.

\paragraph{Equivalence After RMSNorm}

% Therefore, the alternative scaled update yields:
% \begin{equation}
%     \boxed{\bm{x}_k = \frac{1}{\sqrt{k+1}} \left( \bm{x}_0 + \sum_{t=1}^{k} \phi_k\!\left(\mathrm{Norm}(\bm{x}_{t-1})\right) \right)}
% \end{equation}
The final output of the Transformer model with RMSNorm is given by:
\begin{equation}
    \text{FinalOutput} = \mathrm{RMSNorm}\!\left(\bm{x}_{k}\right) = \mathrm{RMSNorm}\!\left(\bm{x}_{k}^{\text{alt}}\right),
\end{equation}
which follows directly from the scale invariance of RMSNorm.
Consequently, instead of analyzing the original dynamics $\bm{x}_k$, we may equivalently consider the rescaled sequence $\bm{x}_k^{\mathrm{alt}}$, which admits the same final output up to a scalar normalization. 

The update scheme in \eqref{altscale} 
 shares the formulation with \eqref{eq3}, with inadaptive angles $\theta = \arccos\left(\frac{\sqrt{k}}{\sqrt{k+1}}\right)$. 
In contrast, the proposed GeoNorm allows the adaptive angle, depending on the step size $\alpha_k$, the current state $\bx_k$, and the update direction $\bs_k$. 
This is convincingly more practical and reliable in manifold optimization. 
We further verify the Pre-Norm and our GeoNorm interpretation and their output equivalence by conducting experiments on comparing their outputs. The following table shows the similar performance of Transformer models using Pre-Norm and Equation \eqref{altscale}, which further supports the equivalence of these two formulations. 

\begin{table}[!htbp]
\caption{The performance of different formulations of Pre-Norm update, with training length 512, Books3 dataset and loss metric. }
\centering
\resizebox{\textwidth}{!}{
\begin{tabular}{cccccccccccc}
\toprule
\textbf{Method} & \textbf{Seed}&   \textbf{5K}  &  \textbf{10K} &   \textbf{15K}     & \textbf{20K} & \textbf{25K} & \textbf{30K} & \textbf{35K} &  
\textbf{40K}  & \textbf{45K}  &  \textbf{50K} \\
\midrule
$x_k+\phi(Norm(x_k))$& 1234&  4.1470 & 3.9028 & 3.8555 & 3.7488 & 3.6815 & 3.6434 & 3.5865 & 3.4906 & 3.4904 & 3.4437\\
$  \frac{\sqrt{k}}{\sqrt{k+1}} \cdot \bm{x}_k + \frac{1}{\sqrt{k+1}} \cdot \phi\!\left(Norm(\bm{x}_k)\right) $ &1234& 	4.1443 & 3.9012 & 3.8533 & 3.7491 & 3.6795 & 3.6454 & 3.5868 & 3.4894 & 3.4898 & 3.4430
\\
\midrule
 $x_k+\phi(Norm(x_k))$& 1235& 4.1618 & 3.9471 & 3.8361 & 3.7779 & 3.6850 & 3.6297 & 3.5751 & 3.5354 & 3.4751 & 3.4548\\
 $  \frac{\sqrt{k}}{\sqrt{k+1}} \cdot \bm{x}_k + \frac{1}{\sqrt{k+1}} \cdot \phi\!\left(Norm(\bm{x}_k)\right) $& 1235&	4.1643 & 3.9500 & 3.8377 & 3.7784 & 3.6885 & 3.6307 & 3.5752 & 3.5347 & 3.4738 & 3.4543
\\
\midrule
 $x_k+\phi(Norm(x_k))$& 1236& 4.1537 & 3.9082 & 3.8357 & 3.7429 & 3.6990 & 3.6268 & 3.5140 & 3.4809 & 3.4542 & 3.4364\\
$  \frac{\sqrt{k}}{\sqrt{k+1}} \cdot \bm{x}_k + \frac{1}{\sqrt{k+1}} \cdot \phi\!\left(Norm(\bm{x}_k)\right) $& 1236&	4.1540 & 3.9070 & 3.8378 & 3.7417 & 3.7002 & 3.6286 & 3.5152 & 3.4816 & 3.4572 & 3.4381
\\
\midrule
$x_k+\phi(Norm(x_k))$& mean& 4.1542 & 3.9194 & 3.8424 & 3.7565 & 3.6885 & 3.6333 & 3.5585 & 3.5023 & 3.4732 & 3.4450 \\
& std & 0.0061 & 0.0197 & 0.0092 & 0.0153 & 0.0076 & 0.0072 & 0.0318 & 0.0237 & 0.0148 & 0.0076 \\
$  \frac{\sqrt{k}}{\sqrt{k+1}} \cdot \bm{x}_k + \frac{1}{\sqrt{k+1}} \cdot \phi\!\left(Norm(\bm{x}_k)\right) $ & mean & 4.1542 & 3.9194 & 3.8429 & 3.7564 & 3.6894 & 3.6349 & 3.5591 & 3.5019 & 3.4736 & 3.4451 \\
& std & 0.0082 & 0.0218 & 0.0073 & 0.0158 & 0.0085 & 0.0075 & 0.0314 & 0.0234 & 0.0133 & 0.0068 \\
\midrule
\end{tabular}
}
\label{table: prenorm}
\end{table}

Taken together, the theoretical analysis and experimental evidence suggest that Pre-Norm can be interpreted as a special case of GeoNorm with a fixed, depth-dependent angle. This perspective also sheds light on the empirical stability advantages of Pre-Norm over Post-Norm. In particular, Pre-Norm implicitly introduces a form of step size decay through its normalized residual structure, whereas Post-Norm lacks such a mechanism. From an optimization perspective, this absence may lead to instability or divergence when step sizes are not carefully controlled, explaining the commonly observed loss spikes and training collapse in Post-Norm Transformers.

% \paragraph{Simplified Representation}

% Let $\hat{\bm{x}}_k = \frac{\bm{x}_k}{\|\bm{x}_k\|}$ and assume $\|\text{Attn}(\hat{\bm{x}}_k)\| = 1$ with $\hat{\bm{x}}_k \perp \text{Attn}(\hat{\bm{x}}_k)$.  Then \eqref{eq:scaled_update} simplifies to:

% \[
% \bm{x}_{k+1} = \sqrt{k} \hat{\bm{x}}_k + \text{Attn}(\hat{\bm{x}}_k),
% \]
% with $\|\bm{x}_{k+1}\| = \sqrt{k+1}$.

% \section{Interpretation as Projected Optimization}

% Pre-Norm can be viewed as an alternating scheme:

% \begin{equation}
%     \begin{aligned}
%         \hat{\bm{x}}_k &= \text{Proj}_{\Omega}(\bm{x}_k) \quad &\text{(project onto sphere)} \\
%         \bm{x}_{k+\frac12} &= \bm{x}_k + \alpha_k \cdot \bm{s}_k(\hat{\bm{x}}_k) \quad &\text{(update in ambient space)} \\
%         \hat{\bm{x}}_{k+\frac12} &= \text{Proj}_{\Omega}(\bm{x}_{k+\frac12}) \quad &\text{(re-project)} \\
%         \bm{x}_{k+1} &= \bm{x}_{k+\frac12} + \beta_k \cdot \bm{t}_k(\hat{\bm{x}}_{k+\frac12}) \quad &\text{(second update)}
%     \end{aligned}
%     \label{eq:pre_norm_opt}
% \end{equation}

% where $\bm{s}_k, \bm{t}_k$ are update directions computed from the projected points, and $\alpha_k, \beta_k$ are effective step sizes.  This differs from Post-Norm, which performs the projection \emph{after} the full update.

\section{The Effect of Clamp Value}
\label{appendix: cost}
\begin{table}[htbp]
\caption{The performance of different clamp values for the implementation in Appendix \ref{appendix: implementation}}
\centering
\resizebox{\textwidth}{!}{
\begin{tabular}{cccccccccccc}
\toprule
\textbf{Model}  &  \textbf{5K}  &  \textbf{10K} &   \textbf{15K}     & \textbf{20K} & \textbf{25K} & \textbf{30K} & \textbf{35K} &  
\textbf{40K}  & \textbf{45K}  &  \textbf{50K} \\
\midrule
$\pi/2$& 4.0650 & 3.8464 & 3.8096 & 3.7072 & 3.6341 & 3.6060 & 3.5471 & 3.4537 & 3.4531 & 3.4092\\
$\pi/4$&		4.0642 & \textbf{3.8423} & \textbf{3.8076} & \textbf{3.7018} & \textbf{3.6317} & \textbf{3.6003} & \textbf{3.5427} & \textbf{3.4486} & \textbf{3.4511} & \textbf{3.4040}\\
$\pi/8$&	\textbf{4.0641} & 3.8501 & 3.8145 & 3.7140 & 3.6471 & 3.6141 & 3.5544 & 3.4638 & 3.4631 & 3.4182
\\
\midrule
\end{tabular}
}
\label{table: length_16384}
\end{table}

\section{Implementation Details}
\label{appendix: implementation}

In this section, we present the implementation of the proposed \methodShort module in \texttt{PyTorch}.

\definecolor{lightgreen}{rgb}{0,0.8,0}
\definecolor{darkgreen}{rgb}{0,0.8,0.2}
\definecolor{backcolour}{rgb}{0.97,0.97,0.94}
\lstset{language=Python,
basicstyle=\smaller\ttfamily,
breaklines=true,
backgroundcolor = \color{backcolour},
keywordstyle=\color{blue}\ttfamily,
stringstyle=\color{lightgreen}\ttfamily,
commentstyle=\color{gray}\ttfamily,
xleftmargin=2.5em,xrightmargin=0.5em, aboveskip=1em,
showstringspaces=False,
morecomment=[l][\color{darkgreen}]{\#}}

\begin{lstlisting}[language=Python]
from tqdm import tqdm
import numpy as np

import torch
import torch.nn as nn

class GeoNorm(nn.Module):
    def __init__(self,clamp=torch.pi/4, method_name="harnomic"):
        super().__init__()
        
        self.scale=nn.Parameter(torch.tensor(1))
        self.bias=nn.Parameter(torch.tensor(0))
        self.clamp=torch.pi/4
        self.method_name=method_name
    def forward(self,x,g,layer_number,layer_total):
        """
        x: residual;
        g: ffn(x) or attn(x);
        layer_number: the current layer index, from 0,1,2, 3 to layer_total-1
        layer_total: layer total number.
        """
        gradient = g - (x * g).sum(dim=-1,keepdim=True) / (
                                            torch.norm(x, p=2, dim=-1, keepdim=True) ** 2) * x

        
        tangent_norm = torch.norm(gradient, p=2, dim=-1, keepdim=True)
        safe_tangent_norm = torch.clamp(tangent_norm, min=1e-8)
        unit_tangent = gradient / safe_tangent_norm

        R = torch.norm(x, p=2, dim=-1, keepdim=True)
        safe_R = torch.clamp(R, min=1e-6)
        theta = torch.clamp(safe_tangent_norm / safe_R, max=self.clamp) 
        if self.method_name=="harmonic":
            theta_attn = torch.clamp((theta* self.scale +self.bias) / (layer_number + 1) , max=self.clamp)
        elif self.method_name=="linear":
            theta = torch.clamp((theta* self.scale +self.bias)*(layer_total-layer_number) / (layer_total) , max=self.clamp)
        elif self.method_name=="sqrt":
            theta = torch.clamp((theta* self.scale +self.bias) / math.sqrt(layer_number + 1) , max=self.clamp)
            
        output = x * torch.cos(theta) + unit_tangent * safe_R * torch.sin(theta)

        return output
\end{lstlisting}

\end{document}